\documentclass{article}

\usepackage{PRIMEarxiv}

\usepackage[utf8]{inputenc} 
\usepackage[T1]{fontenc}    
\usepackage{hyperref}       
\usepackage{url}            
\usepackage{booktabs}       
\usepackage{amsfonts}       
\usepackage{nicefrac}       
\usepackage{microtype}      
\usepackage{lipsum}
\usepackage{fancyhdr}       
\usepackage{graphicx}       
\usepackage{amsmath} 
\usepackage{amssymb}
\usepackage{array} 
\usepackage{multirow} 

\graphicspath{{media/}}     

\usepackage{xcolor}
\definecolor{new}{rgb}{0, 0, 0} %

\title{Causal Learning for Heterogeneous Subgroups Based on Nonlinear Causal Kernel Clustering
}

\author{
  Lu Liu\\
  East China University of Science and Technology \\
  Shanghai, China\\
  \texttt{Y30221662@mail.ecust.edu.cn} \\
   \And
  Yang Tang \\
  East China University of Science and Technology \\
  Shanghai, China\\
  \texttt{yangtang@ecust.edu.cn} \\
   \And
  Kexuan Zhang \\
  East China University of Science and Technology \\
  Shanghai, China\\
  \texttt{Y20220069@mail.ecust.edu.cn} \\
   \And
  Qiyu Sun \\
  East China University of Science and Technology \\
  Shanghai, China\\
  \texttt{Y20220069@mail.ecust.edu.cn}
}

\begin{document}
\maketitle

\begin{abstract}
Due to the challenge posed by multi-source and heterogeneous data collected from diverse environments, causal relationships among features can exhibit variations influenced by different time spans, regions, or strategies. This diversity makes a single causal model inadequate for accurately representing complex causal relationships in all observational data, a crucial consideration in causal learning. To address this challenge, the nonlinear Causal Kernel Clustering method is introduced for heterogeneous subgroup causal learning, highlighting variations in causal relationships across diverse subgroups. \textcolor{new}{The main component for clustering heterogeneous subgroups lies in the construction of the $u$-centered sample mapping function with the property of unbiased estimation, which assesses the differences in potential nonlinear causal relationships in various samples and supported by causal identifiability theory.} Experimental results indicate that the method performs well in identifying heterogeneous subgroups and enhancing causal learning, leading to a reduction in prediction error.
\end{abstract}

\keywords{Causal Learning \and Causal Clustering \and Heterogeneous Subgroups \and Kernel Function}

\section{Introduction}
With the presence of multi-source and heterogeneous data in real-world scenarios, correctly identifying causal relationships from observational data becomes a significant challenge \cite{r1,r2,others2}. However, it is important to recognize that a variety of potential factors can influence causal relationships \cite{others2,others5,r3,r4}. Therefore, addressing heterogeneous subgroups is crucial in causal learning \cite{r6,r19,r20,ker2,r7}. For example, patients may exhibit different, or even opposing, responses (effects) to the same treatment (cause), influenced by unmeasured factors such as nutritional and health status \cite{r7}. This highlights the importance of categorizing patients into distinct subgroups based on their varying treatment responses (causal effects) and developing tailored treatment plans. Within each subgroup, patient responses to treatment should be as consistent as possible, while significant variations are expected across different subgroups.

\textcolor{new}{Various methods have been developed for heterogeneous subgroup causal learning \cite{ker2,r17,r10,r11,r13}, enabling the detection of variations in causal relationships across different subgroups. The emphasis of these methods varies widely, with focusing on data types \cite{r10,r11}, specific models \cite{r13}, and task-specific scenarios \cite{r14,r15}, respectively. Additionally, some methods impose extra conditions \cite{s1} to ensure their theoretical operability. While these methods perform effectively within their intended contexts and demonstrate impressive accuracy, their applicability may be limited outside those specific conditions. Rather than imposing specific distributional assumptions, \cite{ker2} assumes the availability of sufficient samples for statistical inference, alongside the standard causal Markov and faithfulness assumptions. Innovatively, they introduce a distance covariance-based kernel and identify clusters that minimize structural heterogeneity, facilitating causal structure learning within the isometric mapping space.}

\textcolor{new}{To better mitigate the overestimation or underestimation of subgroups information, a $u$-centered method \cite{lemma1} is introduced to the sample mapping function to ensure unbiased estimation. This matrix ensures unbiased inner product estimation, facilitating distance covariance estimation even with small sample sizes. Additionally, \cite{r17} highlights the causal clustering procedure as a means to enhance causal learning. Based on these, a plug-and-play module is developed here, seamlessly integrating with existing causal learning approaches. This flexible module leverages subgroup information to enhance model's comprehension of diversity in causal relationships, addressing the challenge of insufficient representation by a single causal model. This flexibility extends its applicability, making it suitable for a broad spectrum of scenarios.}

The main contributions of this paper are as follows:
\begin{itemize}
\item \textcolor{new}{A $u$-centered sample mapping function is introduced, designed to map samples into a high-dimensional space isomorphic to the causal graph while ensuring unbiased estimation. This function enables reliable estimation, thereby enhancing the accuracy and effectiveness of subsequent causal clustering operations.}
\item As a flexible causal discovery module, the method seamlessly integrates with existing causal learning approaches and has been validated in both climate modeling and house price prediction. Experimental results demonstrate its effectiveness in enhancing causal learning performance.
\end{itemize}

\section{Related Work}
There is significant attention given to tackling the challenge of heterogeneous subgroup causal learning with distinct focuses. For known heterogeneous subgroups, data fusion is proposed in \cite{r21} to extract unique insights from combined data, enabling a more comprehensive understanding than individual datasets can provide. Meta-learning is introduced in causal inference in \cite{r22} to focus on tackling covariate bias by alleviating distributional biases in both training and test data. Furthermore, the method for estimating heterogeneous treatment effects is also proposed in  \cite{r23} to customize personalized treatments based on insights from diverse subgroups.

\textcolor{new}{Identifying heterogeneous subgroups in real-world scenarios is often impractical, as these subgroups typically emerge only through the analysis of observational data. Given the practical constraints that hinder the anticipation of such subgroups in advance, identifying heterogeneous subgroups from observational data becomes a significant challenge. Various methods have been proposed to address this issue \cite{ker2,r10,r11}. \textcolor{new}{\cite{ker2} introduces a distance covariance-based kernel to identify clusters and subsequently performs causal structure learning within each identified heterogeneous subgroup.} Considering the influence of the temporal dimension on causal relationships, the ACD algorithm in \cite{r10} incorporates dynamic information sharing. Furthermore, \cite{r11} proposes dual-process intervention learning to handle non-stationary time-series data, facilitating the mutual reinforcement of domain indicator recovery and causal structure learning.}

\section{Problem Description}
Given a set of samples, the focus is on heterogeneous subgroup causal learning \cite{ker2,r17,r11,r13}, aiming to classify samples based on causal relationships. Within each subgroup, samples are expected to exhibit similar nonlinear causal interactions among features, while causal relationships may differ across subgroups.

\begin{figure*}[t]
        \centering
    \includegraphics[width=14cm]{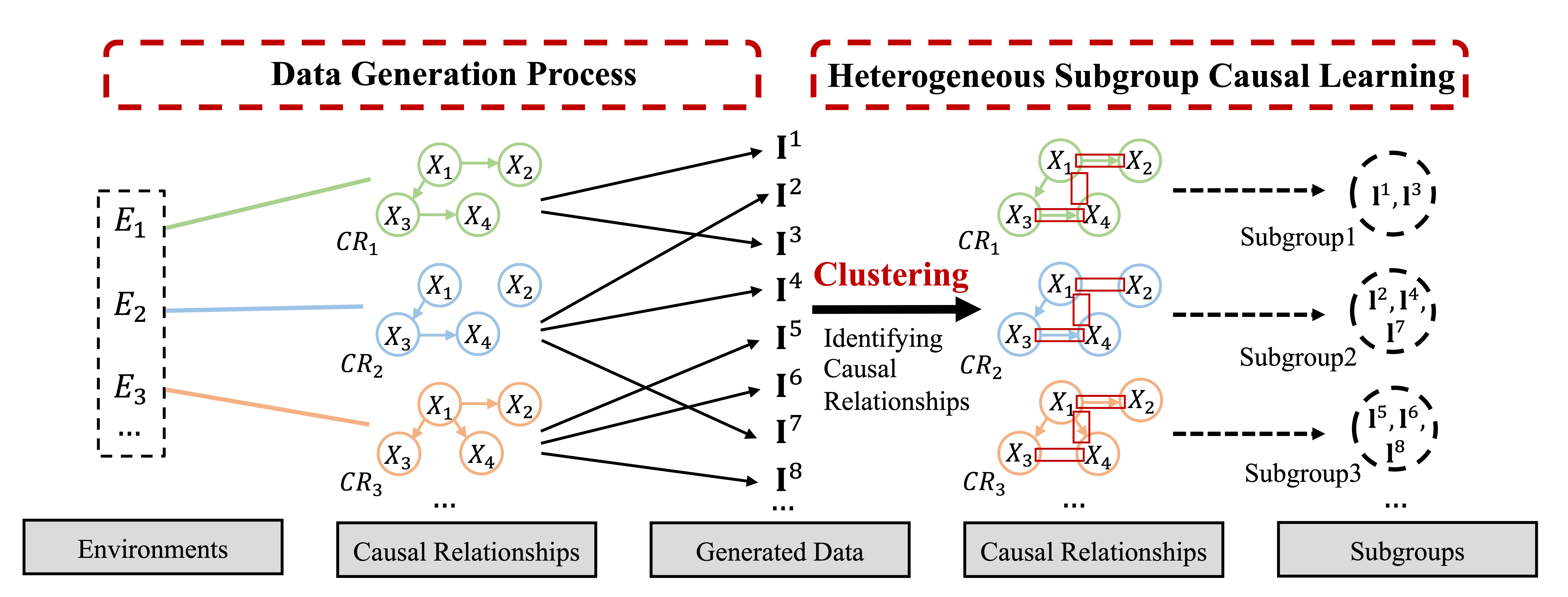}
    \caption{An illustration of the process of generating multi-source and heterogeneous data from various environments and the approach to heterogeneous subgroup causal learning, taking four features ($m=4$) as an example. $E_{1},\dots$ denote the multiple environments, $X_{1},\dots$ denote the features, $CR_{1},\dots$ denote the diverse causal relationships between features, and $\textbf{l}^{s} (s\in \{1,2,\dots\})$ denote the data generated by the corresponding causal relationships. The objective is to obtain heterogeneous subgroups through clustering, identifying various causal relationships.}
\end{figure*}

Suppose the $m$ features from $s$-th sample, denoted as $\textbf{l}^{s}=[l_{1}^s,\dots,l_{m}^s]\in \Bbb{R}^{m}$. The values of these features adhere to the causal generation process described by:
\begin{equation}\label{1}
    l_{k}^s=\text{f}_{k}^{s}(\text{pa}(k,s))+e_{k}^s,
\end{equation}
for $k\in \{1,\dots,m\}$. In \eqref{1}, $\text{pa}(k,s)$ represents the direct causal parent features of $l_{k}^s$, influencing $l_{k}^s$ casually. The nonlinear term $\text{f}_{k}^{s}$ represents the causal relationships between $\text{pa}(k,s)$ and $l_{k}^s$. The term $e_{k}^s$ introduces a non-Gaussian noise component, accounting for some potential unmeasured factors.

Notably, the variation in causal relationships across subgroups is greater than that within subgroups, leading to the following definition of subgroup invariance.

\newtheorem{definition}{Definition}[section]
\begin{definition}[Subgroup Invariance]\label{Def3.1}
For any feature $X_{k}, k\in \{1,\dots,m\}$ and the subgroups of samples $\text{G}=\{\text{g}_{1},\dots,\text{g}_{r}\}$, there exists subgroup invariance regarding the causal relationships, which corresponds to \eqref{1}:
\begin{equation}\label{2}
    X_{k} = \text{f}_{k}^{\text{g}_{i}}(\text{pa}(k,\text{g}_{i}))+e_{k}^{\text{g}_{i}}.
\end{equation}
\end{definition}

Drawing from Definition \ref{Def3.1}, the nonlinear \underline{\textbf{C}}ausal \underline{\textbf{K}}ernel \underline{\textbf{C}}lustering (CKC) method is introduced. The method aims at clustering samples with similar nonlinear causal relationships and identifying heterogeneous subgroups, thereby revealing variations in causal relationships across different subgroups  \cite{ker2,r17,r13}.

\section{Method}
Numerous studies have deeply explored the field of heterogeneous subgroup causal learning  \cite{ker2,r17,r11,r13}.  \cite{r17} employs kernelized heterogeneous risk minimization, which achieves the dual objectives of heterogeneous subgroup exploration and causal invariant learning in the kernel space, offering a new insight for nonlinear scenarios. In terms of heterogeneous subgroup exploration, it constructs a kernel clustering algorithm based on the sample mapping function, enabling precise capture of heterogeneous subgroup information, which is crucial for causal invariant learning. \textcolor{new}{Inspired by \cite{ker2,r17,lemma1}, a causal clustering method with a $u$-centered based sample mapping function is proposed to capture heterogeneous subgroup information. The mapping function captures nonlinear relationship within samples and ensures unbiased estimation. By capturing heterogeneous subgroup information through the clustering operation, the information is leveraged for causal signal warning and stable feature mining, thereby enhancing the effectiveness and reliability of subsequent causal learning methods.}

\textcolor{new}{To ensure that the sample mapping function effectively captures heterogeneous subgroup information, it is crucial to consider both the dependence and independence between features during the construction process \cite{r2,r15,s1,DAG1}. In statistics, dependence is typically measured using correlation coefficients \cite{others2,lemma1,r27,r28,lemma2}, while independence is often verified using the Chi-square test \cite{ker2,tj1,tj2,tj3}. Therefore, a binary decision method based on dependence/independence is considered, where the correlation coefficient statistic is used to accurately assess dependencies between feature pairs \cite{lemma1,r28,lemma2}, and the Chi-square statistic is employed to test feature independence \cite{ker2,tj1,tj2,tj3}. By comparing dependence \cite{lemma1,r28,lemma2} and independence \cite{tj1,tj2,tj3}, the $u$-centered sample mapping function is built based on a binary decision method, which represents the relationships within the data and facilitates the capture of complex patterns.}
\textcolor{new}{To summarize, a $u$-centered sample mapping function is introduced to center the samples, map them into a high-dimensional space, and ensure unbiased estimation.}

\subsection{Marginal Distance Covariance}

In this section, the metric is designed to measure the dependence between features. In statistics, correlation coefficients are commonly used to describe linear dependence between features \cite{others2,r27,corr2}. For nonlinear dependence, more complex statistics are required, such as distance correlation coefficients \cite{others2,ker2,lemma1,r28,lemma2}. The $u$-centered version of distance matrix \cite{lemma1,discorr2} is proposed to construct an extended form of the concept of distance covariance \textcolor{new}{for unbiased estimation}. Through eliminating sample bias, \cite{lemma1} results in a standardized coefficient for evaluating nonlinear dependence between features. Inspired by \cite{lemma1,r28,discorr2}, \textcolor{new}{the concept of marginal distance covariance is considered} based on the $u$-centered version of distance matrix \textcolor{new}{with the property of unbiased estimation}.

\textbf{Notation.} In this paper, both $n$ and $n'$ represent the number of samples, $m$ represents the number of features. For any $d$-dimensional square matrix $\textbf{M}=(m_{st})_{s, t = 1}^{d} \in \Bbb{R}^{d\times d}$, $\tilde{\textbf{M}}=(\tilde{m}_{st})_{s, t = 1}^{d} \in\Bbb{R}^{d\times d}$ is used to represent the $u$-centered version \cite{lemma1,discorr2} of matrix $\textbf{M}$, where $\tilde{m}_{st}=m_{st}-\frac{1}{d-2}\sum_{u=1}^{d}m_{ut}-\frac{1}{d-2}\sum_{v=1}^{d}m_{sv}+\frac{1}{(d-1)(d-2)}\sum_{u,v=1}^{d}m_{uv}$. For any matrix $\textbf{N}\in \Bbb{R}^{d_{1}\times d_{2}}$, let $\textbf{N}_{i,}$ and $\textbf{N}_{,j}$ represent the $i$-th row and the $j$-th column in $\textbf{N}$, respectively.
\newtheorem{lemma}{Lemma}[section]
\begin{lemma}[Sample Distance Matrix \cite{ker2,lemma1,lemma2,discorr2}]\label{lemma4.1}
    For samples $\textbf{S}=[X_{1},\dots,X_{m}]\in \Bbb{R}^{n\times m}$, the sample distance matrix is denoted as $\textbf{H} \in \Bbb{R}^{n\times n\times m}$, with $ \textbf{H}_{i,i',j}=|\textbf{S}_{i,j}-\textbf{S}_{i',j}|$. $\textbf{H}_{\cdot,\cdot,j}$ denotes the sample distance matrix for feature $X_{j}$.
\end{lemma}

 \cite{lemma1} employs the $u$-centered version to design the partial distance covariance for ternary features. Since the study focuses on the causal relationships between all pairs of features, \textcolor{new}{the marginal distance covariance is considered} by using the $u$-centered version \cite{lemma1,r28,discorr2}, allowing for a more appropriate evaluation of the dependence between pairs of features.

\begin{definition}[Marginal Distance Covariance \cite{lemma1,r28,discorr2}]\label{def4.1}
For two features $X_{p}$ and $X_{q}$, with their sample distance matrices $\textbf{H}_{\cdot,\cdot,p}=[\mathcal{P}_{1},\cdots,\mathcal{P}_{n}]$ and $\textbf{H}_{\cdot,\cdot,q}=[\mathcal{Q}_{1},\cdots,\mathcal{Q}_{n}]$, the marginal distance covariance between the features is defined as:

\begin{equation}\label{3}
\text{mdCov}^{2}(X_{p},X_{q})=\sum_{\alpha=1}^{n}\sum_{\beta=1}^{n}\text{dCov}^{2}(\mathcal{P}_{\alpha},\mathcal{Q}_{\beta}),
\end{equation}

\begin{equation}\label{4}
    \text{dCov}^{2}(\mathcal{P}_{\alpha},\mathcal{Q}_{\beta})=\int_{\Bbb{R}^{2n}} \frac{|\phi_{\mathcal{P}_{\alpha},\mathcal{Q}_{\beta}}(\delta,\rho)-\phi_{\mathcal{P}_{\alpha}}(\delta)\phi_{\mathcal{Q}_{\beta}}(\rho)|^{2}}{c|\delta|^{n+1}|\rho|^{n+1}}d\delta d\rho,
\end{equation}
where $\mathcal{P}_{\alpha}$, $\mathcal{Q}_{\beta}$ represent the corresponding low-dimensional components of the sample distance matrices $\textbf{H}_{\cdot,\cdot,p}$ and $\textbf{H}_{\cdot,\cdot,q}$, $c=\pi^{n+1}/\tau^{2}(\frac{n+1}{2})$, $|\cdot|$ represents the Euclidean norm defined as $|x|=\sqrt{\overline{x}^{T}x}$ for and vector $x$ ($\overline{x}$ represents the conjugate of $x$), $\phi_{\mathcal{P}_{\alpha}}$ and $\phi_{\mathcal{Q}_{\beta}}$ are the characteristic functions of $\mathcal{P}_{\alpha}$ and $\mathcal{Q}_{\beta}$ respectively, $\phi_{\mathcal{P}_{\alpha},\mathcal{Q}_{\beta}}$ is the joint characteristic function. The alternative simplified form of $\text{dCov}^{2}(\mathcal{P}_{\alpha},\mathcal{Q}_{\beta})$ is given as:

\begin{equation}\label{5}
    \text{dCov}^{2}(\mathcal{P}_{\alpha},\mathcal{Q}_{\beta})=<\tilde{\textbf{A}}_{\alpha},\tilde{\textbf{B}}_{\beta}>,
\end{equation}
where $\textbf{A}_{\alpha}$, $\textbf{B}_{\beta}$ represent the sample distance matrices generated from $\mathcal{P}_{\alpha}$, $\mathcal{Q}_{\beta}$, and $\tilde{\textbf{A}}_{\alpha}$, $\tilde{\textbf{B}}_{\beta}$ represent the $u$-centered versions of $\textbf{A}_{\alpha}$, $\textbf{B}_{\beta}$.
\end{definition}

By eliminating sample bias from the marginal distance covariance \cite{lemma1}, \textcolor{new}{the corresponding marginal distance correlation coefficient can be obtained,} which serves as a metric to evaluate the nonlinear dependence between two features \cite{others2,lemma1,r28}.

\begin{equation}\label{6}
    \text{mdCor}^{2}(X_{p},X_{q})=\frac{\text{mdCov}^{2}(X_{p},X_{q})}{\sqrt{\text{mdCov}^{2}(X_{p},X_{p})}\cdot\sqrt{\text{mdCov}^{2}(X_{q},X_{q})}}.
\end{equation}

\textit{Remark}. Current methods \cite{r27,linear2,linear1} predominantly focus on quantifying linear causal relationships. To extend this capability, the marginal distance covariance is introduced as a metric for capturing nonlinear causal relationships.

\subsection{Sample Mapping Function Based on Hypothesis Testing}
In this section, \textcolor{new}{the construction of metric is explored} for determining the independence between pairs of features and \textcolor{new}{the sample mapping function is considered to} reflect the corresponding causal relationships between features. As is well known, the Chi-squared statistic is widely used for testing the independence of features \cite{ker2,tj1,tj2}. For instance, in medical research, Chi-squared test can be used to examine whether a particular disease is independent of different lifestyle factors such as smoking or drinking. In market research, it can be used to analyze whether consumers preference for different brands are independent of factors such as gender or age. Thus, the hypothesis framework \textcolor{new}{is considered} for independence testing and \textcolor{new}{the Chi-squared statistic is used} as an indicator to assess the independence between features \cite{r28,lemma2,tj2}.

\begin{lemma}[Hypothesis Testing \cite{r28,lemma2,tj2}]\label{lemma4.2}
    The hypothesis testing is given to ascertain the independence of two features:
\begin{equation}\label{7}
    \text{H}_{0}=\{(X_{p},X_{q}) | X_{p} \perp \!\!\! \perp
 X_{q}\},
\end{equation}
\begin{equation}\label{8}
    \text{H}_{1}=\{(X_{p},X_{q}) | X_{p} \not \! \perp \!\!\! \perp
 X_{q}\}.
\end{equation}
If the condition $\text{H}_{0}$ holds, $X_{p}$ and $X_{q}$ are independent. Conversely, if the condition $\text{H}_{1}$ holds, there exists nonlinear dependence between the two features.
\end{lemma}

Expanding upon Lemma \ref{lemma4.2}, \textcolor{new}{the sample mapping function is considered, utilizing marginal distance covariance to gauge the similarities among samples.} Specifically, \textcolor{new}{the marginal distance covariance is seen through Definition \ref{def4.1}}. After removing the sample bias, it can be used to measure the dependence between features \cite{lemma1,r28,lemma2}. Meanwhile, utilizing the independence hypothesis test from Lemma \ref{lemma4.2}, \textcolor{new}{the common Chi-squared statistics \cite{ker2,tj1,tj2} can be used} to assess the independence between features. By leveraging the disparity between these two statistics, the sample mapping function can assess both dependence and independence between features, providing the representation of the underlying causal relationships.

\begin{definition}[Sample Mapping Function \cite{ker2,r17,lemma1,r28,tj2}]\label{def4.2}
    For samples $\textbf{S} \in \Bbb{R}^{n\times m}$, consider the sample distance matrix $\textbf{H} \in \Bbb{R}^{n\times n\times m}$, and its $u$-centered processed normalized matrix $\textbf{Z} \in \Bbb{R}^{n\times n\times m}$, $\textbf{Z}_{i,i',j}=\frac{\textbf{C}_{i,i',j}}{\overline{\textbf{H}}_{\cdot,\cdot,j}}$, where $\textbf{C}_{\cdot,\cdot,j}$ is the $u$-centered version of $\textbf{H}_{\cdot,\cdot,j}$ and $\overline{\textbf{H}}_{\cdot,\cdot,j}$ is the mean over $\textbf{H}_{\cdot,\cdot,j}$. Define the sample mapping function $\Phi(\textbf{S}_{i,})$: $\Bbb{R}^{m}\to \Bbb{R}^{m\times m}$ as:
\begin{equation}\label{9}
\Phi(\textbf{S}_{i,})=\sum_{\alpha=1}^{n}\sum_{\beta=1}^{n}\sum_{\zeta=1}^{n}V_{\zeta,\alpha}\cdot V_{\zeta,\beta}^{T}-\Gamma(\nu),
\end{equation}
\begin{equation}\label{10}
    V_{\zeta,\gamma}=\text{Vec} (|\textbf{Z}_{\zeta,\gamma,\cdot}-\textbf{Z}_{i,\gamma,\cdot}|), \forall{\gamma}\in \{\alpha,\beta\},
\end{equation}
\begin{equation}\label{11}
    \Gamma(\nu)_{j,j'}=
    \begin{cases}
        n\chi_{1-\nu}^{2}(1), & j \neq j'\\
        0, & j=j'
    \end{cases},
\end{equation}
for $i\in \{1,2,\dots,n\}$, where $|\cdot|$ represents the absolute value, $\text{Vec}$ represents the column vector form, $\Gamma(\nu)\in \Bbb{R}^{m\times m}$, $\nu$ represents the significance level, and $\chi_{1-\nu}^{2}(1)$ represents the Chi-square value with 1 degree of freedom.
\end{definition}

\textit{Proof}. The first term of Definition \ref{def4.2} represents the form of $u$-centered based marginal distance covariance after eliminating bias: $\sum_{\alpha=1}^{n}\sum_{\beta=1}^{n}\sum_{\zeta=1}^{n}V_{\zeta,\alpha}\cdot V_{\zeta,\beta}^{T}$, which is the marginal distance correlation coefficient that can evaluate feature dependence \cite{lemma1,r28,lemma2}. Compared to the conventional distance correlation coefficient, it incorporates an extra layer of summation operation. This results in an $n$-fold amplification of the second term Chi-squared statistic $n\chi_{1-\nu}^{2}(1)$ with the degree of freedom for a single sample being 1 to ensure the accuracy of the result \cite{tj3,freedom1,freedom2}. From the perspective of the sample mapping function $\Phi(\textbf{S}_{i,})$, the element in the $p$-th row and $q$-th column represents the disparity between the marginal distance correlation coefficient statistic and the Chi-squared statistic for features $X_{p}$ and $X_{q}$. If the element is greater than 0, it indicates that the non-linear dependence of features reflected by the marginal distance correlation coefficient statistic is more significant than the independence of features reflected by the Chi-square statistic: $(\sum_{\alpha=1}^{n}\sum_{\beta=1}^{n}\sum_{\zeta=1}^{n}V_{\zeta,\alpha}\cdot V_{\zeta,\beta}^{T})_{p,q} > (n\chi_{1-\nu}^{2}(1))_{p,q}$, i.e., $(\sum_{\alpha=1}^{n}\sum_{\beta=1}^{n}\sum_{\zeta=1}^{n}V_{\zeta,\alpha}\cdot V_{\zeta,\beta}^{T})_{p,q} - (n\chi_{1-\nu}^{2}(1))_{p,q} >0$. It suggests a stronger nonlinear dependence between features. Conversely, if the element is less than 0: $(\sum_{\alpha=1}^{n}\sum_{\beta=1}^{n}\sum_{\zeta=1}^{n}V_{\zeta,\alpha}\cdot V_{\zeta,\beta}^{T})_{p,q} \leq (n\chi_{1-\nu}^{2}(1))_{p,q}$, i.e., $(\sum_{\alpha=1}^{n}\sum_{\beta=1}^{n}\sum_{\zeta=1}^{n}V_{\zeta,\alpha}\cdot V_{\zeta,\beta}^{T})_{p,q} - (n\chi_{1-\nu}^{2}(1))_{p,q} \leq 0$, the features are more in line with the independence. \(\blacksquare\)

\textit{Remark}. Differing from the conventional emphasis on causal relationships between features, the sample mapping function shifts the focus to the samples themselves, exploring both dependence and independence between features through a binary decision method. This approach enables a nuanced and comprehensive understanding of the causal relationships between features.

\newtheorem{theorem}{Theorem}[section]
\begin{theorem}\label{thm4.1}
	For a significance level $\nu$, concerning features $X_{p}$ and $X_{q}$, if the condition $\sum_{i=1}^{n}\Phi(\textbf{S}_{i,})_{p,q}>0$ holds, it implies that the features are not independent i.e., $X_{p} \not \! \perp \!\!\! \perp X_{q}$, indicating the presence of nonlinear dependence. Conversely, if the condition $\sum_{i=1}^{n}\Phi(\textbf{S}_{i,})_{p,q}\leq 0$ holds, it implies that the features are independent i.e., $X_{p} \perp \!\!\! \perp X_{q}$.
\end{theorem} 
\textit{Proof}. As is shown above, for features $X_{p}$ and $X_{q}$, if $(\sum_{\alpha=1}^{n}\sum_{\beta=1}^{n}\sum_{\zeta=1}^{n}V_{\zeta,\alpha}\cdot V_{\zeta,\beta}^{T})_{p,q} > (n\chi_{1-\nu}^{2}(1))_{p,q}$, i.e., $(\sum_{\alpha=1}^{n}\sum_{\beta=1}^{n}\sum_{\zeta=1}^{n}V_{\zeta,\alpha}\cdot V_{\zeta,\beta}^{T})_{p,q} - (n\chi_{1-\nu}^{2}(1))_{p,q} >0$, it indicates that the degree of nonlinear dependence between features is more significant than the degree of feature independence. \textcolor{new}{The conclusion is that}: $X_{p} \not \! \perp \!\!\! \perp X_{q}$. Conversely, if $(\sum_{\alpha=1}^{n}\sum_{\beta=1}^{n}\sum_{\zeta=1}^{n}V_{\zeta,\alpha}\cdot V_{\zeta,\beta}^{T})_{p,q} \leq (n\chi_{1-\nu}^{2}(1))_{p,q}$, i.e., $(\sum_{\alpha=1}^{n}\sum_{\beta=1}^{n}\sum_{\zeta=1}^{n}V_{\zeta,\alpha}\cdot V_{\zeta,\beta}^{T})_{p,q} - (n\chi_{1-\nu}^{2}(1))_{p,q} \leq 0$, it indicates that the features are more in line with the independence. \textcolor{new}{The conclusion is that}: $X_{p} \perp \!\!\! \perp X_{q}$. Meanwhile, considering the potential presence of outliers in multi-source and heterogeneous data, \cite{r27} emphasizes that these outliers can lead to incorrect final decisions. Thus, \textcolor{new}{the summation form is seen} to make the final binary decision: $\sum_{i=1}^{n}\Phi(\textbf{S}_{i,})_{p,q}>0$ for $X_{p} \not \! \perp \!\!\! \perp X_{q}$, and $\sum_{i=1}^{n}\Phi(\textbf{S}_{i,})_{p,q}\leq0$ for $X_{p} \perp \!\!\! \perp X_{q}$. \(\blacksquare\)

\textit{Remark}. \textcolor{new}{Dependence between features has been assessed by examining the difference between distance covariance and critical values \cite{ker2}. The Theorem \ref{thm4.1} follows a binary decision-making approach based on dependence and independence, quantifying the degree of dependence and independence between features separately and making the final decision by comparing the strengths of these two aspects.}

The subsequent section explores the causal identifiability theory of the sample mapping function $\Phi(\textbf{S}_{i,})$, providing support for Theorem \ref{thm4.1} and the clustering.

\subsection{Nonlinear Causal Kernel}
Given the primary use of mapping to assess sample similarity rather than directly inferring causal relationships between features, the corresponding nonlinear causal kernel is constructed for the sample mapping function \cite{ker2,r17,ker3}.
Inspired by \cite{ker2,r17}, the nonlinear causal kernel \textcolor{new}{is constructed} to measure the similarity of samples in the mapping space, ensuring the clustering more accurately. \cite{ker2,lemma1} define a cosine-similarity measure to assess similarity of samples in the mapped space, with a combination of inner products and norms. \cite{ker2,s2,Fcausal1,F1} propose Frobenius inner-product and norm for theoretical design in causal learning. Building upon these methods \cite{ker2,r17,lemma1,s2,Fcausal1}, the nonlinear causal  kernel \textcolor{new}{is considered} for the sample mappinng function through cosine-similarity from and Frobenius inner-product norm combinations, providing an approximate formula from the perspective of inner-product for subsequent proof.

\begin{definition}[Nonlinear Causal Kernel \cite{ker2,r17,lemma1,s2}]\label{def4.3}
    Integrating the sample mapping function $\Phi(\textbf{S}_{i,})$, the Frobenius inner product $<\cdot,\cdot>_{\text{F}}$, and its norm $\|\cdot\|_{\text{F}}$, define the nonlinear causal kernel $\kappa(\textbf{S}_{i,},\textbf{S}_{i',})$ between two samples as:
\begin{equation}
\kappa(\textbf{S}_{i,},\textbf{S}_{i',})=\frac{<\Phi(\textbf{S}_{i,}),\Phi(\textbf{S}_{i',})>_{\text{F}}}{\| \Phi(\textbf{S}_{i,})\|_{\text{F}}\cdot \| \Phi(\textbf{S}_{i',})\|_{\text{F}}}.
\end{equation}
\end{definition}

The nonlinear causal kernel is constructed through: $\kappa(\textbf{S}_{i,},\textbf{S}_{i',}) \approx cosin(\theta) = \frac{<A, B>}{|A|\cdot|B|}=\frac{<A, B>}{\sqrt{<A, A>}\cdot \sqrt{<B, B>}}=\frac{<\Phi(\textbf{S}_{i,}),\Phi(\textbf{S}_{i',})>_{\text{F}}}{\sqrt{<\Phi(\textbf{S}_{i,}),\Phi(\textbf{S}_{i,})>_{\text{F}}}\cdot \sqrt{<\Phi(\textbf{S}_{i',}),\Phi(\textbf{S}_{i',})>_{\text{F}}}}=\frac{<\Phi(\textbf{S}_{i,}),\Phi(\textbf{S}_{i',})>_{\text{F}}}{\| \Phi(\textbf{S}_{i,})\|_{\text{F}}\cdot \| \Phi(\textbf{S}_{i',})\|_{\text{F}}}\propto <\Phi(\textbf{S}_{i,}),\Phi(\textbf{S}_{i',})>_{\text{F}} = \sum_{j=1}^{m}\sum_{j'=1}^{m}\Phi(\textbf{S}_{i,})_{j,j'}\cdot \Phi(\textbf{S}_{i',})_{j,j'}$. Thus, it can be approximated by the formula: \(\kappa(\textbf{S}_{i,},\textbf{S}_{i',})\propto \sum_{j=1}^{m}\sum_{j'=1}^{m}\Phi(\textbf{S}_{i,})_{j,j'}\cdot \Phi(\textbf{S}_{i',})_{j,j'}\).

\begin{figure}[t]
        \centering
    \includegraphics[width=12cm]{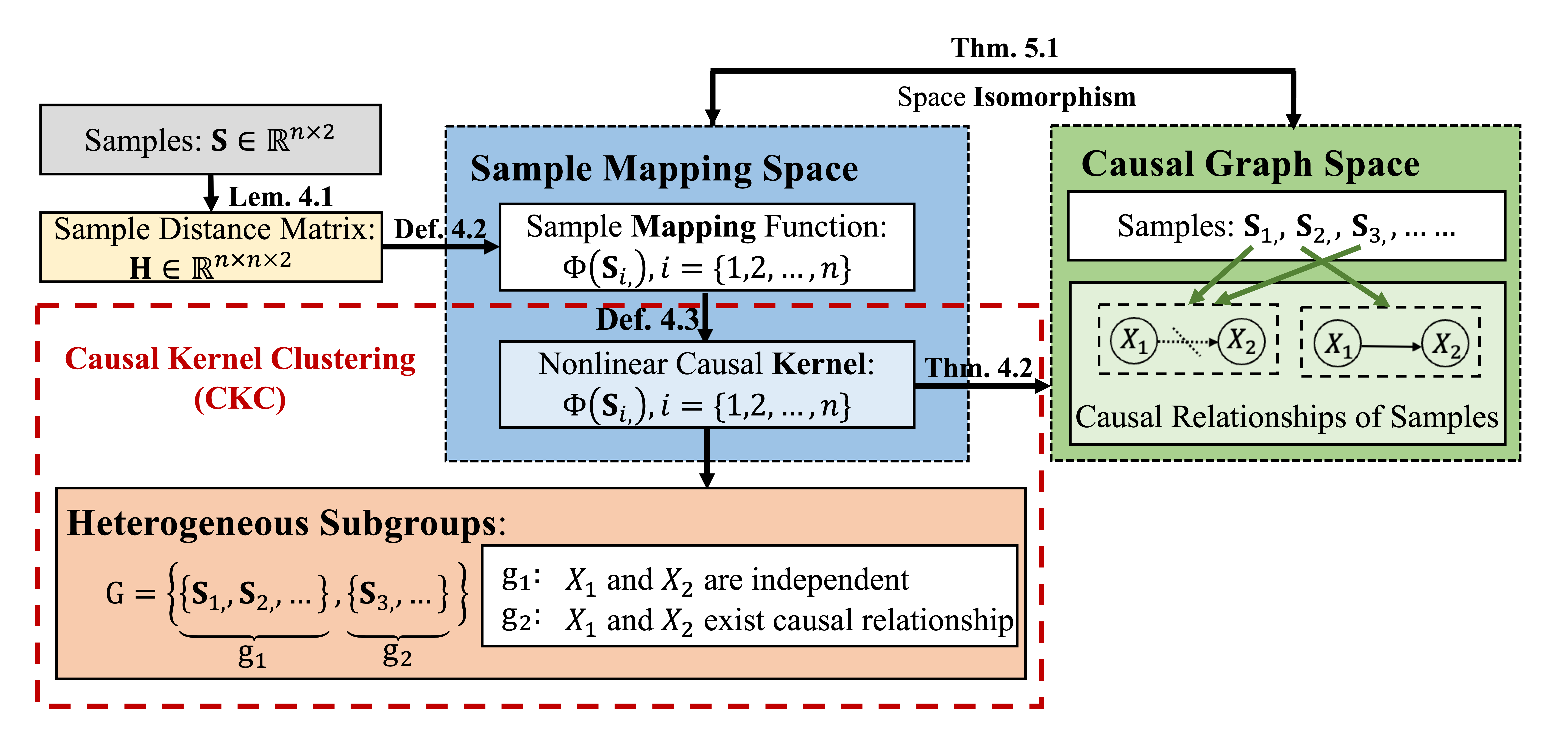}
    \caption{The operated framework of the method (CKC) which is based on samples $\textbf{S} \in \Bbb{R}^{n\times 2}$ with two features ($m=2$).}
    \label{fig1}
\end{figure}

\textcolor{new}{The kernel clustering method based on the sample mapping function will be delved into next. It will be proven that, by leveraging Theorem \ref{thm4.2}, this method can unearth heterogeneous subgroup causal information, which in turn will enhance the performance of subsequent causal learning methods.}

\begin{theorem}\label{thm4.2}
Consider two sample sets $\textbf{S}\in \Bbb{R}^{n\times m}$ and $\textbf{S}'\in \Bbb{R}^{n'\times m}$ with the same features $X = (X_{1},\dots,X_{m})$. If the condition $\sum_{i=1}^{n}\sum_{i'=1}^{n'}\kappa(\textbf{S}_{i,},\textbf{S}'_{i',})<0$ holds, it indicates the presence of distinct nonlinear causal relationships between the two sample sets.
\end{theorem}

\textit{Proof}. \textcolor{new}{Based on Definition \ref{def4.3}, if the condition $\sum_{i=1}^{n}\sum_{i'=1}^{n'}\kappa(\textbf{S}_{i,},\textbf{S}'_{i',})<0$ holds, it is evident that at least one pair of features exists, denoted as $(j,j')\in \{1,\dots,m\}$, where $\Phi(\textbf{S}_{i,})_{j,j'}$ and $\Phi(\textbf{S}'_{i',})_{j,j'}$ exhibit opposite signs. Assume $\Phi(\textbf{S}_{i,})_{j,j'}>0$ and $\Phi(\textbf{S}'_{i',})_{j,j'}<0$. According to Theorem \ref{thm4.1}, it can be proved that $X_{j} \not \! \perp \!\!\! \perp X_{j'}$ in samples $\textbf{S}$, but $X_{j} \perp \!\!\! \perp X_{j'}$ in samples $\textbf{S}'$. It implies that there exists different nonlinear causal relationships underpinning two sample sets $\textbf{S}$ and $\textbf{S}'$. Conversely, the same reasoning holds. \(\blacksquare\)}

\textit{Remark}. \textcolor{new}{The consistency of decision criteria in causal inference has been established using Slutsky’s Theorem and the continuous mapping theorem, as demonstrated through the \(\mathcal{I}\)-function framework \cite{ker2}. The Theorem \ref{thm4.2} is directly derived from Definition \ref{def4.3} and Theorem \ref{thm4.1}, offering a intuitive perspective. Drawing on these theoretical foundations, the direction of elements in the sample mapping function \(\Phi(\textbf{S}_{i,})\) reveals variations in causal relationships among features. Theorem \ref{thm5.1} further proves that the nonlinear causal kernel, constructed upon the sample mapping function, can distinguish different causal relationships between features. This capability enables the method to extract heterogeneous subgroup causal information, thereby enhancing the performance of subsequent causal learning methods.}

Based on the theoretical framework presented earlier, \textcolor{new}{the nonlinear CKC method is depicted in Fig. \ref{fig1}} with illustration employing two features $X_{1}$ and $X_{2}$ ($m=2$). For samples $\textbf{S} \in \Bbb{R}^{n\times 2}$, the distance matrix $\textbf{H}\in \Bbb{R}^{n\times n\times 2}$ is initially constructed. Subsequently, the sample mapping function $\Phi(\textbf{S}_{i,})$ is employed to map the samples to a high-dimensional space, and the kernel function $\kappa(\textbf{S}_{i,},\textbf{S}_{i',})$ is then constructed, where $i,i'\in \{1,2,\dots,n\}$. In the subsequent section, \textcolor{new}{it is established theoretically that the sample mapping space is isomorphic to the causal graph space, which ensures that the subgroups identified by the method are causally heterogeneous.} As shown in Fig. \ref{fig1}, a notable example illustrates the absence of causal relationship between features in both samples $\textbf{S}_{1,}$ and $\textbf{S}_{3,}$, while such a causal relationship exists in sample $\textbf{S}_{2,}$. Considering the different causal relationships between features, the samples $\textbf{S}_{1,}$ and $\textbf{S}_{3,}$ are more likely clustered into one subgroup.

\section{Causal Identifiability Theory}

\subsection{Space Isomorphism}

\textcolor{new}{To demonstrate that the method can identify heterogeneous subgroups and be applied in causal learning methods, it is necessary to consider two aspects. On one hand, the rationality of the constructed mapping function needs to be ensured, indicating that the operations in the mapping space are consistent with those in the original sample space.} On the other hand, it needs to explore the causal identifiability of the sample mapping function through mapping space, ensuring that the method could reveal and quantify the causal relationships directly.

Initially, it is necessary to demonstrate that the sample mapping function $\Phi(\textbf{S}_{i,})$ is bijective, indicating isomorphism between the original sample space $\Bbb{R}^{m}$ and the mapping space $\Bbb{R}^{m\times m}$ with a property of distance invariance \cite{others5,tonggou1,tonggou2}.

The bijective property of a mapping function is frequently utilized to establish an isomorphism between the mapping space and the original space in mathematics, ensuring that the mapping space and the original space share the same properties \cite{others5,ker2,tonggou1,tonggou2}. Therefore, to ensure the rationality and effectiveness of using the mapping space instead of the original space for the subsequent clustering operation, it is necessary to prove the bijective property of the sample mapping function firstly. Based on the commonly used methods for proving bijective property in the field of mathematics through injective property and surjective property, the sample mapping function is proved by separately proving these two properties \cite{ys2,ys3}.

\textit{Proof}. Consider any mapping function $\Phi(\textbf{S}_{i,})\in \Bbb{R}^{m\times m}$ and $\Phi(\textbf{S}_{i',})\in \Bbb{R}^{m\times m}$, if the condition $\Phi(\textbf{S}_{i,})=\Phi(\textbf{S}_{i',})$ holds, it implies that $\textbf{Z}_{i,\cdot,\cdot}=\textbf{Z}_{i',\cdot,\cdot}$ as well as $\textbf{S}_{i,}=\textbf{S}_{i',}$, where $\textbf{S}_{i,}\in \Bbb{R}^{m}$ and $\textbf{S}_{i',}\in \Bbb{R}^{m}$. This ensures that the sample mapping function $\Phi(\textbf{S}_{i,}): \Bbb{R}^{m}\to \Bbb{R}^{m\times m}$ is injective; Furthermore, the mapping space $\Bbb{R}^{m\times m}$ is  comprised of the sample mapping function $\Phi(\textbf{S}_{i,})$. For each element $\eta\in \Bbb{R}^{m\times m}$, there exists a sample $\textbf{S}_{\epsilon,}$ such that $\Phi(\textbf{S}_{\epsilon,})=\eta$, confirming that the sample mapping function $\Phi(\textbf{S}_{i,})$: $\Bbb{R}^{m}\to \Bbb{R}^{m\times m}$ is surjection. \(\blacksquare\)

\subsection{Causal Identifiability}
This section demonstrates the causal identifiability of the developed sample mapping function $\Phi(\textbf{S}_{i,\cdot})$.
\textcolor{new}{To demonstrate that the method can identify heterogeneous subgroups and be applied in causal learning methods, the causal identifiability of the sample mapping function in the mapping space needs to be explored, showing the relationship between the mapping space and the causal graph space. It is common in mathematics to establish theoretical analyses by defining two spaces and proving their relationships \cite{ker2,tonggou1,tonggou3}. Therefore, the concept of two spaces needs to be defined. Inspired by \cite{others5,r6,ker2,others1}, the concept of equivalent causal graph with certain conditions needs to be designed, which further demonstrates that the mapping function can reflect causal meanings. To this end, a new concept of $m$-connectivity is considered:}
$N^{\mathcal{G'}}$.
\begin{definition}[Causal Graph Space \cite{others5,r6,ker2,others1}]\label{def5.1}
    Consider a causal graph $\mathcal{G}=(O^{\mathcal{G}},E^{\mathcal{G}})$, where $O^{\mathcal{G}}$ represents the nodes of the causal graph $\mathcal{G}$, and $E^{\mathcal{G}}$ represents the edges of the causal graph $\mathcal{G}$. $N^{\mathcal{G}}$ represents the pairs of nodes $(j,j')$ with $m$-connectivity, indicating that the longest path between the nodes is $m$. For two causal graphs $\mathcal{G}=(O^{\mathcal{G}},E^{\mathcal{G}})$ and $\mathcal{G'}=(O^{\mathcal{G'}},E^{\mathcal{G'}})$ with $O^{\mathcal{G}}=O^{\mathcal{G'}}$, they are equivalent if and only if $N^{\mathcal{G}}=N^{\mathcal{G'}}$. The $\Bbb{G}^{m}$ is used to represent the space of equivalent causal graphs.
\end{definition}
\textcolor{new}{A simple example and a mathematical proof are presented below to illustrate the validity of Definition \ref{def5.1}.}

\textit{Example.} Since the longest path of a node does not exceed the total number of nodes in the causal graph, for two causal graphs $\mathcal{G}=(O^{\mathcal{G}},E^{\mathcal{G}})$ and $\mathcal{G'}=(O^{\mathcal{G'}},E^{\mathcal{G'}})$ with same four nodes: $O^{\mathcal{G}}=O^{\mathcal{G'}}=\{O_1,O_2,O_3,O_4\}$, the longest path $m$ only makes sense when taking 1, 2, or 3.
Assume that causal graph $\mathcal{G}$ is: $E^{\mathcal{G}}=O_2 - O_1 - O_3 - O_4$, and causal graph $\mathcal{G'}$ is: $E^{\mathcal{G'}}=O_1 - O_2 - O_4 - O_3$, having different causal relationships, the \( m \)-connectivity set of causal graph $\mathcal{G}$ with the longest path lengths of 1, 2, and 3:

\begin{center}
$N^{\mathcal{G}}=
    \begin{cases}
        \{(O_1,O_2),(O_1,O_3),(O_3,O_4)\}, & m=1\\
        \{(O_1,O_4),(O_2,O_3)\}, & m=2\\
        \{(O_2,O_4)\}, & m=3.
    \end{cases}$
\end{center}

The \( m \)-connectivity set of causal graph $\mathcal{G'}$ with the longest path lengths of 1, 2, and 3:

\begin{center}
$N^{\mathcal{G'}}=
    \begin{cases}
        \{(O_1,O_2),(O_2,O_4),(O_3,O_4)\}, & m=1\\
        \{(O_1,O_4),(O_2,O_3)\}, & m=2\\
        \{(O_1,O_3)\}, & m=3.
    \end{cases}$
\end{center}

The different \( m \)-connectivity set $N^{\mathcal{G}} \neq N^{\mathcal{G'}}$ indicates that $\mathcal{G}=(O^{\mathcal{G}},E^{\mathcal{G}})$ and $\mathcal{G'}=(O^{\mathcal{G'}},E^{\mathcal{G'}})$ are not equivalent and belong to the different causal relationship. Therefore, the \( m \)-connectivity property of node pairs can be used as a criterion for determining the equivalence of causal graphs.

\textcolor{new}{Next, the causal matrix space needs to be explored from the perspective of the sample mapping function $\Phi(\textbf{S}_{i,\cdot})$. Specifically, the design of an equivalent square matrix that the sample mapping function maps to under certain conditions needs to be examined \cite{r6,ker2,others1},} laying the foundation for demonstrating that the mapping function can capture causal relationships. \cite{DAG1,s2,others1,matrix1} introduce the concept of the adjacency, which represents the edge weights in a causal graph. \cite{s2} proposes using the adjacency matrix to guide network layers in capturing complex causal relationships between features. Inspired by \cite{ker2,DAG1,s2,others1,matrix1}, \textcolor{new}{the notion of causal matrix is seen to} define equivalent condition for the sample mapping function $\Phi(\textbf{S}_{i,\cdot})$. From the meaning behind Definition \ref{def4.2}, it can be seen that the direction of the sample mapping function elements is crucial for determining the causal relationship between features. If the positive or negative direction of elements in two sets of samples match, it means the causal relationship between features is the same. Therefore, \textcolor{new}{the causal matrix is constructed} to formalize the direction of the sample mapping function results $\Phi(\textbf{S}_{i,\cdot})$ and used as an equivalent condition. \textcolor{new}{In the causal matrix, 1 and -1 are used to represent the different directions: 1 for positive direction and -1 for negative direction.}

\begin{definition}[Causal Matrix Space \cite{ker2,s2,others1,matrix1}]\label{def5.2}
    Consider $m$-order square matrix $\textbf{Y}$ and its causal matrix $\text{sign}(\textbf{Y}) \in \Bbb{R}^{m \times m}$ which is defined in \eqref{10}. For two $m$-order square matrices $\textbf{Y}$ and $\textbf{Y}'$, they are equivalent if and only if $\text{sign}(\textbf{Y})=\text{sign}(\textbf{Y}')$. The $\Bbb{Y}^{m}$ is used to represent the space of equivalent causal matrices.
\begin{equation}
    \text{sign}(\textbf{Y})_{j,j'}=
    \begin{cases}
        1, & \textbf{Y}_{j,j'}\geq 0\\
        -1, & \textbf{Y}_{j,j'} < 0
    \end{cases}.
    \label{13}
\end{equation}
\end{definition}

\textit{Remark}. \textcolor{new}{The causal matrix and its associated space can formalize the potential causal relationships represented by the sample mapping function $\Phi(\textbf{S}_{i,\cdot})$. By building on the concept of orthant equivalence, \cite{ker2} generalizes the concept of quadrants from \(R^2\) to higher dimensions. To clarify this concept, there is a simple example along with a mathematical proof for illustration.}

\textit{Example}. Suppose the space of square matrix mapped by the sample mapping function $\Phi(\textbf{S}_{i,\cdot})$ is two-dimensional: $\Phi(\textbf{S}_{i,\cdot}) \in \Bbb{R}^{2\times 2}$. According to Definition \ref{def4.2} and Definition \ref{def5.2}, for two square matrices mapped by the sample mapping function: \(\Phi(\textbf{S}_{i,\cdot})=\begin{bmatrix} a_{11} & a_{12} \\ a_{21} & a_{22} \end{bmatrix}=\begin{bmatrix} 0 & 0.8 \\ -0.2 & 0 \end{bmatrix}\), and \(\Phi(\textbf{S}_{i',\cdot})=\begin{bmatrix} a'_{11} & a'_{12} \\ a'_{21} & a'_{22} \end{bmatrix}==\begin{bmatrix} 0 & 0.2 \\ -0.5 & 0 \end{bmatrix}\), their respective causal matrices are denoted as: \(\text{sign}(\Phi(\textbf{S}_{i,\cdot}))=\begin{bmatrix} 1 & 1 \\ -1 & 1 \end{bmatrix}\) and \(\text{sign}(\Phi(\textbf{S}_{i',\cdot}))=\begin{bmatrix} 1 & 1 \\ -1 & 1 \end{bmatrix}\). \textcolor{new}{It can be seen that} $\text{sign}(\Phi(\textbf{S}_{i,\cdot}))$ and $\text{sign}(\Phi(\textbf{S}_{i',\cdot}))$ are the same. For the construction and meaning of the sample mapping function $\Phi(\textbf{S}_{i,\cdot})$, the consistence of the element directions implies that the causal relationships underlying the features are consistent. Therefore, the causal matrix can serve as a criterion for determining the equivalence of square matrices. Meanwhile, it could reflect the underlying causal relationships captured by the sample mapping function.

Finally, \textcolor{new}{the causal identifiability of the sample mapping function is explored}, ensuring that the method can identify heterogeneous subgroups and integrate with established causal learning methods.

\begin{theorem}\label{thm5.1}
    The sample mapping function $\Phi(\textbf{S}_{i,})$ is an isomorphic mapping from the causal graph space $\Bbb{Y}^{m}$ to the causal matrix space $\Bbb{G}^{m}$, thereby ensuring the isomorphism between the sample mapping space and the causal graph space.
\end{theorem}

\textcolor{new}{There is an intuitive understanding for the Theorem \ref{thm5.1} above.}

\textcolor{new}{\textit{Proof}. For two causal graphs behind the data: $\mathcal{G}$ and $\mathcal{G'}$, according to Definition \ref{def5.1}, suppose they share the same pairs of nodes $m$-connectivity: $N^{\mathcal{G}}=N^{\mathcal{G'}}$. It implies that $\mathcal{G}$ and $\mathcal{G'}$ exhibit the same causal relationship: $\mathcal{G}=\mathcal{G'}$. Let 
\begin{equation}
    \textbf{I}_{j,j'}=
    \begin{cases}
        1, & (j,j')\in N^{\mathcal{G}}\\
        -1, & (j,j')\notin N^{\mathcal{G}}
    \end{cases},
    \label{14}
\end{equation}
the matrix $\textbf{I}_{j,j'}$ can be constructed based on the $m$-connectivity set of node pair to represents the same meaning as \eqref{13} in Definition \ref{def5.2}. Specifically, when node pair $(j,j')$ belongs to this $m$-connectivity set, it indicates that there exists a causal relationship between nodes $(j,j')$, denoted as 1. Conversely, when node pair $(j,j')$ does not belong to the $m$-connectivity set, it signifies the absence of the causal relationship, denoted as -1. For the square matrix mapped by the sample mapping function $\Phi(\textbf{S}_{i,})$, if the corresponding matrix element is positive: $\Phi(\textbf{S}_{i,})_{j,j'}>0$, it indicates a causal relationship between the features $X_{j}$ and $X_{j'}$: $X_{j} \perp \!\!\! \perp X_{j'}$ . If the corresponding matrix element is negative: $\Phi(\textbf{S}_{i,})_{j,j'}<0$, it signifies no causal relationship between $X_{j}$ and $X_{j'}$: $X_{j} \not \! \perp \!\!\! \perp X_{j'}$. Thus, in the causal graph space, if the node pair $(j,j')$ belongs to the $m$-connectivity set, it is equivalent to the matrix element corresponding to features $X_j$ and $X_j'$ being positive. \(\blacksquare\)}

\textit{Remark}. \textcolor{new}{The Theorem \ref{thm5.1} revolves around the sample mapping function $\Phi(\textbf{S}_{i,})$ in Definition \ref{def4.2} to prove its validity, ensuring that the sample mapping function is consistent with the causal graph space in expressing causal relationships. The existence of causal relationships is directly associated with the matrix, which indicates the direction of the elements in the sample mapping function $\Phi(\textbf{S}_{i,})$. This association provides a intuitive understanding of the causal implications embodied by the mapping function $\Phi(\textbf{S}_{i,})$. Meanwhile, existing works on equivalence classes of ancestral graphs and orthant equivalence classes of symmetric real matrices have explored group isomorphisms as a means to facilitate subsequent causal inference \cite{ker2}.}

These definitions and theorems contribute to a understanding of the causal relationships revealed within the sample mapping function and its corresponding high-dimensional representation.

\section{Experiments}
In this section, experiments are conducted to validate the method from two perspectives. Both the generated synthetic data and real-world data sourced from the Indian Ocean Dipole \cite{r29} are employed to assess the accuracy of the method in identifying heterogeneous subgroups by identifying causal relationships between features. Furthermore, it is also proposed to deploy the method with popular causal learning methods to validate its effectiveness in enhancing causal learning using real-world Boston Housing Data from Kaggle\label{website}.
\footnote{https://www.kaggle.com/datasets/vikrishnan/boston-house-prices.\label{website}}

\subsection{Synthetic Data}

\textit{Baseline and Evaluation Metrics.} \textcolor{new}{Various clustering methods are employed as benchmarks for performance comparison. K-means clustering (K-means) \cite{r30}, a well-known partitional clustering method without a kernel, assigns samples into k clusters based on their proximity to cluster centers. Polynomial kernel clustering (Poly) \cite{r31}, designed to capture nonlinear relationships, maps samples into a higher-dimensional space to identify complex class boundaries. Radial basis function kernel clustering (RBF)\cite{r32}, particularly effective for clustering nonlinearly separable samples, corresponds to an infinite-dimensional feature space, allowing for highly flexible decision boundaries. To assess the performance of these methods, both the V-measure Score ($\text{V-measure}$) and the Adjusted Rand Index ($\text{ARI}$) are employed as evaluation metrics.}
\begin{center}
    $\text{V-measure} = \frac{2 \times \text{H} \times \text{C}}{\text{H}+\text{C}}$,
\end{center}
\begin{center}
    $\text{ARI} = \frac{\text{RI}-\text{E}(\text{RI})}{\text{Max}(\text{RI})-\text{E}(\text{RI})}$,
\end{center}
where $\text{H}$ represents the homogeneity of the cluster, $\text{C}$ represents the completeness of the cluster, and $\text{RI}$ represents the Rand Index.

\textit{Experimental setup.} Six random Directed Acyclic Graphs (DAGs) \cite{s5} are used to simulate heterogeneous subgroups, created with 10 variables \( X = (X_1, \dots, X_{10}) \), to model the causal relationships between them. In linear scenarios, random data with zero correlation (\( \text{corr} = 0 \)) between features is first generated by the process: 

\[
\mathbf{SD} = \left[ SD_{X_1}, \dots, SD_{X_{10}} \right] \in \mathbb{R}^{100 \times 10}, \mathbf{SD} = \sigma \cdot \mathbf{F} + \mathbf{\mu},
\]
where \( \sigma \) follows a uniform distribution \( U(0.5, 2) \), \( \mathbf{F} \in \mathbb{R}^{100 \times 10} \) is a matrix with elements from a normal distribution \( N(0, 1) \), and \( \mathbf{\mu} \in \mathbb{R}^{100} \) is a column vector with elements from a uniform distribution \( U(-4, 4) \). Subsequently, random data with non-zero correlation (\( \text{corr} \neq 0 \)) is generated by the following process:

\[
SD_{X_p} = SD_{X_p} + \sum_{X_q \in \text{pa}(p)} w_{p,q} \cdot SD_{X_q},
\]

where \( \text{pa}(p) \) represents the direct causal parent variables of \( X_p \), and \( w_{p,q} \) represents the weight between \( X_p \) and \( X_q \). In nonlinear scenarios, data are generated by adding randomized Gaussian noise.

\begin{table*}[h]
\scriptsize
\renewcommand\arraystretch{1.25}
\centering
\caption{The clustering results of methods on both linear and nonlinear scenarios, where the higher $\text{V-measure}$ and $\text{ARI}$ denote better performances.}

\begin{tabular}{|c|cccccccc|cc|}
\hline
\multirow{2}{*}{} & \multicolumn{8}{c|}{Linear} & \multicolumn{2}{c|}{Nonlinear} \\ \cline{2-9}
                  & \multicolumn{2}{c|}{$\text{corr}\neq0$, $\mu \sim U(-4,4)$}                                                  & \multicolumn{2}{c|}{$\text{corr}\neq0$, $\mu=0$}                                                  & \multicolumn{2}{c|}{$\text{corr}=0$, $\mu \sim U(-4,4)$}                                                  & \multicolumn{2}{c|}{$\text{corr}=0$, $\mu=0$}                             & \multicolumn{2}{c|}{}                              \\ \hline
Methods           & \multicolumn{1}{c|}{$\text{V-measure}$}     & \multicolumn{1}{c|}{$\text{ARI}$}           & \multicolumn{1}{c|}{$\text{V-measure}$}     & \multicolumn{1}{c|}{$\text{ARI}$}           & \multicolumn{1}{c|}{$\text{V-measure}$}     & \multicolumn{1}{c|}{$\text{ARI}$}           & \multicolumn{1}{c|}{$\text{V-measure}$}     & $\text{ARI}$           & \multicolumn{1}{c|}{$\text{V-measure}$}     & $\text{ARI}$           \\ \hline
K-means \cite{r30}           & \multicolumn{1}{c|}{0.59}          & \multicolumn{1}{c|}{0.53}          & \multicolumn{1}{c|}{0.52}          & \multicolumn{1}{c|}{0.35}          & \multicolumn{1}{c|}{0.55} & \multicolumn{1}{c|}{0.38} & \multicolumn{1}{c|}{0.51}          & 0.31          & \multicolumn{1}{c|}{0.07}          & 0.14          \\ \hline
Poly \cite{r31}             & \multicolumn{1}{c|}{0.68}          & \multicolumn{1}{c|}{0.65}          & \multicolumn{1}{c|}{0.53}          & \multicolumn{1}{c|}{0.38}          & \multicolumn{1}{c|}{0.62}          & \multicolumn{1}{c|}{0.44}          & \multicolumn{1}{c|}{0.48}          & 0.25          & \multicolumn{1}{c|}{0.14}          & 0.15          \\ \hline
RBF \cite{r32}               & \multicolumn{1}{c|}{0.65}          & \multicolumn{1}{c|}{0.59}          & \multicolumn{1}{c|}{0.56}          & \multicolumn{1}{c|}{0.47}          & \multicolumn{1}{c|}{0.61}          & \multicolumn{1}{c|}{0.42}          & \multicolumn{1}{c|}{0.14}          & 0.21          & \multicolumn{1}{c|}{0.26}          & 0.15          \\ \hline
DEP-CON \cite{ker2}           & \multicolumn{1}{c|}{\textbf{0.85}}          & \multicolumn{1}{c|}{\textbf{0.77}}          & \multicolumn{1}{c|}{0.37}          & \multicolumn{1}{c|}{0.23}          & \multicolumn{1}{c|}{\textbf{0.71}} & \multicolumn{1}{c|}{\textbf{0.57}}          & \multicolumn{1}{c|}{0.52}          &     0.33      & \multicolumn{1}{c|}{0.15}          &  0.08         \\ \hline
\textbf{CKC}      & \multicolumn{1}{c|}{0.79} & \multicolumn{1}{c|}{0.71} & \multicolumn{1}{c|}{\textbf{0.65}} & \multicolumn{1}{c|}{\textbf{0.56}} & \multicolumn{1}{c|}{0.65}          & \multicolumn{1}{c|}{0.47}          & \multicolumn{1}{c|}{\textbf{0.59}} & \textbf{0.43} & \multicolumn{1}{c|}{\textbf{0.40}} & \textbf{0.23} \\ \hline
\end{tabular}
\label{tab1}
\end{table*}

\textit{Results.} As shown in TABLE \ref{tab1}, the method outperforms others in nonlinear scenarios, demonstrating its ability to handle data with nonlinear causal relationships between features and highlighting its broad applicability in real-world settings. \textcolor{new}{In linear scenario, the method performs better than \cite{r30,r31,r32}. Compared to \cite{ker2}, the method remains susceptible to noise interference and exhibits lower performance in linear scenarios with noise. Specifically, \cite{ker2} relies solely on multiplication in its mapping process, which results in less noise propagation and better noise filtering capability, contributing to stronger robustness. The method here employs a multi-layer summation operation. While exploring the complex structure and distribution of data, it also amplifies the noise during propagation, increasing its sensitivity to noise. Therefore, further enhancements are needed to improve performance in linear scenarios with noise.}


\subsection{Real-world Data}
In this section, the accuracy of the method in identifying heterogeneous subgroups is initially examined. Furthermore, the method is integrated into popular causal learning methods to assess the effectiveness of heterogeneous subgroup information in enhancing causal learning.

\subsubsection{Experiments on Indian Ocean Dipole Data}
\textit{Baseline and Evaluation Metrics.} Three methods are used for comparisons for IOD warning: the National Centers for Environmental Prediction coupled forecast system model version 2 (CFSv2), the European Centre for Medium-Range Weather Forecasts seasonal forecast system version 5 (ECs5), and the Network-based Approach (Net-based) \cite{r29}. The following metrics Accuracy, Recall, and F1-score are used as the metrics to illustrate the prediction accuracy, allowing for a comprehensive assessment of the ability to accurately identify IOD events:
\begin{center}
    $\text{Accuracy} = \frac{\text{TP} + \text{TN}}{\text{TP} + \text{TN} + \text{FP} +\text{FN}}$,
\end{center}
\begin{center}
    $\text{Recall} = \frac{\text{TP}}{\text{TP} + \text{FN}}$,
\end{center}
\begin{center}
$\text{F1-score} = \frac{2\times \text{TP}}{\text{FP}+\text{FN}}$,
\end{center}
where $\text{TP}$ represents true positive value, $\text{TN}$ represents true negative value, $\text{FP}$ represents false positive value, and $\text{FN}$ represents false negative value.

\begin{figure}[t]
        \centering
    \includegraphics[width=10cm]{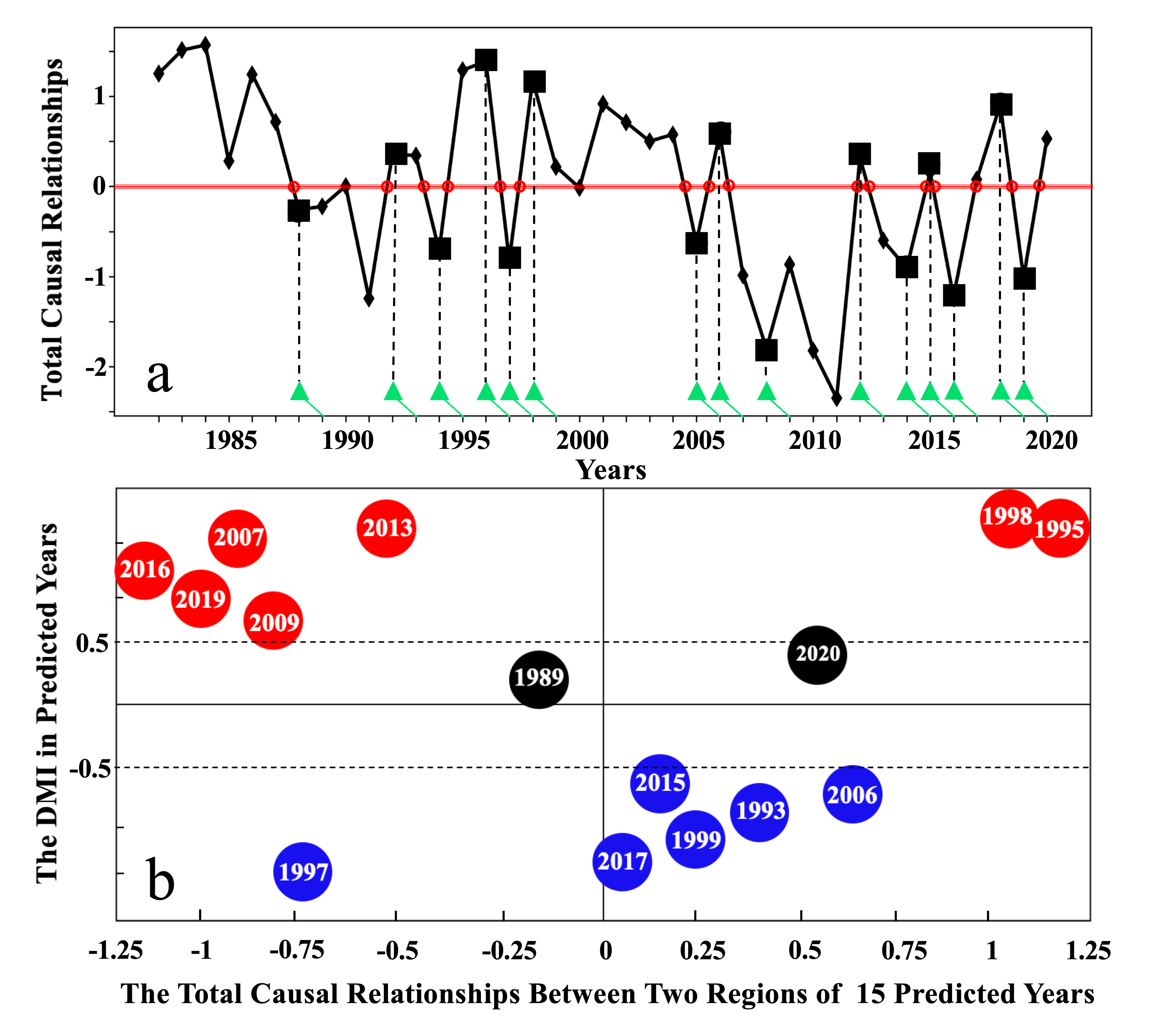}
    \caption{(a) The evolution of the standardized $\text{YC}(y)$ time series. When it undergoes transitions (passing through red line) and reaches an extreme value (marked by black boxes), there is a possibility of an IOD event in the upcoming year (pointed with the green line). (b) The years predicted by the method are displayed in the circles. Both circles of blue and red represent IOD events among the correctly predicted. The black circles represent the wrongly predicted.}
    \label{fig2}
\end{figure}

\textit{Experimental setup.} The public time-series data from  \cite{r29} are employed for analyzing the Indian Ocean Dipole (IOD) event across two regions: the western region ($10^{\circ}$N to $10^{\circ}$S, $50^{\circ}$E to $70^{\circ}$E) and the eastern region (0 to $10^{\circ}$S, $90^{\circ}$E to $110^{\circ}$E). The western region is partitioned into 441 grid nodes and the eastern region into 231 grid nodes, representing different areas. The analysis involves daily data spanning from 1982 to 2020, treating the nodes as features. To enable the classification of heterogeneous subgroups, the sample mapping function is devised, mapping the sample of each time point to a high-dimensional space.

The nonlinear causal kernel \cite{ker2,r17,lemma1,s2} $\kappa_{i,j}^{t}$ is constructed as the causal relationships between nodes by calculating the values of the sample mapping function at intersection times for each pair of nodes across the two regions, using a time point $t$ of 60 days and a time lag parameter $\theta$ set between 0 and 100 days :

\begin{equation}
    \kappa_{i,j}^{t}(\theta)=\frac{<\Phi(T_{i}(t-\theta)),\Phi(T_{j}(t))>_{\text{F}}}{\| \Phi(T_{i}(t-\theta))\|_{\text{F}}\cdot\| \Phi(T_{j}(t))\|_{\text{F}}},
\end{equation}
\begin{equation}
    \kappa_{i,j}^{t}(-\theta)=\frac{<\Phi(T_{i}(t)),\Phi(T_{j}(t-\theta))>_{\text{F}}}{\| \Phi(T_{i}(t))\|_{\text{F}}\cdot \| \Phi(T_{j}(t-\theta))\|_{\text{F}}},
\end{equation}
where $<\cdot,\cdot>_{\text{F}}$ represents the Frobenius inner product, and $\| \cdot\|_{\text{F}}$ represents its norm, with $i$ and $j$ representing the nodes from regions of west ($\text{WE}$) and east ($\text{EA}$), respectively.

\begin{table*}[h]
\renewcommand\arraystretch{1.25}
\centering
\scriptsize
\caption{The elements of the confusion matrices and the evaluation of methods mentioned in  \cite{r29}. The direction of the arrow represents the direction in which the evaluation metrics are expected to vary.}
\begin{tabular}{|c|c|c|c|c|c|c|c|}
\hline
Methods                & True Positive ($\uparrow$)          & True Negative ($\uparrow$)          & False Positive ($\downarrow$)         & False Negative ($\downarrow$)         & $\text{Accuracy}$ ($\uparrow$)      & $\text{Recall}$ ($\uparrow$)        & $\text{F1-score}$ ($\uparrow$)      \\ \hline
CFSv2\cite{r29}                  & 4           & 16          & 11         & 5          & 0.56          & 0.44          & 0.33          \\ \hline
ECs5\cite{r29}                   & 6           & 18          & 6          & 4          & 0.71          & 0.60          & 0.55          \\ \hline
Net-based\cite{r29} & 11          & 16          & 4          & 6          & 0.73          & 0.65          & 0.69          \\ \hline
\textbf{CKC}          & \textbf{13} & \textbf{22} & \textbf{2} & \textbf{2} & \textbf{0.90} & \textbf{0.87} & \textbf{0.87} \\ \hline
\end{tabular}
\label{tab2}
\end{table*}

The total causal relationship $\text{TC}(t)$ between the two regions at a given time point $t$ is the sum of the causal relationships of all node pairs:
\begin{equation}
    \text{TC}(t)=\sum_{i\in \text{WE}, j \in \text{EA}}\kappa_{i,j}^{t}(\theta)+\kappa_{i,j}^{t}(-\theta).
\end{equation}

Since the causal relationships between regions differ between years with IOD events and normal years \cite{s4}, the $\text{YC}(y)$ is designed as the total causal relationships between two regions in a given year $y$, denoted as $\text{YC}(y)=\sum_{t\in y} \text{TC}(t), y \in \{1982,\dots,2020\}$, allowing exploration of early causal warning signals.

\textit{Results.} Fig. \ref{fig2} (a) illustrates the evolution of the standardized $\text{YC}(y)$ time series. \textcolor{new}{The time series YC(y) is proposed to be utilized to cluster years into two subgroups, corresponding to years with IOD events and normal years.} The results of clustering are determined based on the signs of $\text{YC}(y)$, indicating different causal relationships between years with $\text{YC}(y)>0$ and $\text{YC}(y)<0$ (heterogeneous subgroups). Thus, it is inferred that when the sign of $\text{YC}(y)$ undergoes a transition and $\text{YC}(y)$ reaches an extreme value, the causal relationships between regions have probably varied. A fascinating finding is that an IOD event is more likely to occur in the following year due to these variations in causal relationships, serving as an early causal warning signal one year in advance. Fig. \ref{fig2} (b) displays the predicted years with DMI \cite{r29}, providing 15 warnings out of 15 IOD events, with 13 of these predictions being accurate. The ability of the method to identify heterogeneous subgroups and provide early causal warning is explicitly demonstrated. TABLE \ref{tab2} shows the details of the confusion matrices and the evaluation metrics for the methods, highlighting the capacity of the method to predict more anomalous years while minimizing misleading predictions. This underscores the accuracy of the method in ensuring the reliability of identifying heterogeneous subgroups.

\subsubsection{Experiments on Boston Housing Data}

\textit{Baseline and Evaluation Metrics.} The method is deployed with popular causal learning methods to showcase the effectiveness of heterogeneous subgroup information in enhancing causal learning: ERM \cite{r16}, for its broad applicability and effectiveness across diverse problem types; KerHRM \cite{r17}, for its robustness to heterogeneity in data distributions; and Stable Learning \cite{DWR,SRDO,SVI}, for its better performance in the presence of data distribution shifts between training and testing phases. The $\text{RMSE}$ and the $\text{Sta\_Error}$ are used as metrics to evaluate these enhancements:
\begin{center}
$\text{RMSE}=\sqrt{\frac{1}{n}\sum_{k=1}^{n}(Y_{k}-\hat{Y}_{k})^{2}}$,
\end{center}
\begin{center}
$\text{Sta\_Error}=\text{RMSE}(D^{\text{test}})$,
\end{center}
where $n$ represents the number of samples, $Y_{k}$ represents the true value of sample $k$, $\hat{Y}_{k}$ represent the predicted result for sample $k$, and $D^{\text{test}}$ represents the test samples.

\textit{Experimental setup.} The ability of the model to distinguish between causal and non-causal factors is essential, ensuring a focus on factors with causal relationships to the target. The method achieves this by exploring the stability of relationships between features and the target using heterogeneous subgroup information. Features exhibiting higher stability of relationships are considered more likely to be causal factors. Subsequently, it is integrated with popular causal learning methods, prompting them to prioritize attention to these causal factors, thereby enhancing their generalizability.

\begin{figure}[t]
\centering
    \includegraphics[width=12cm]{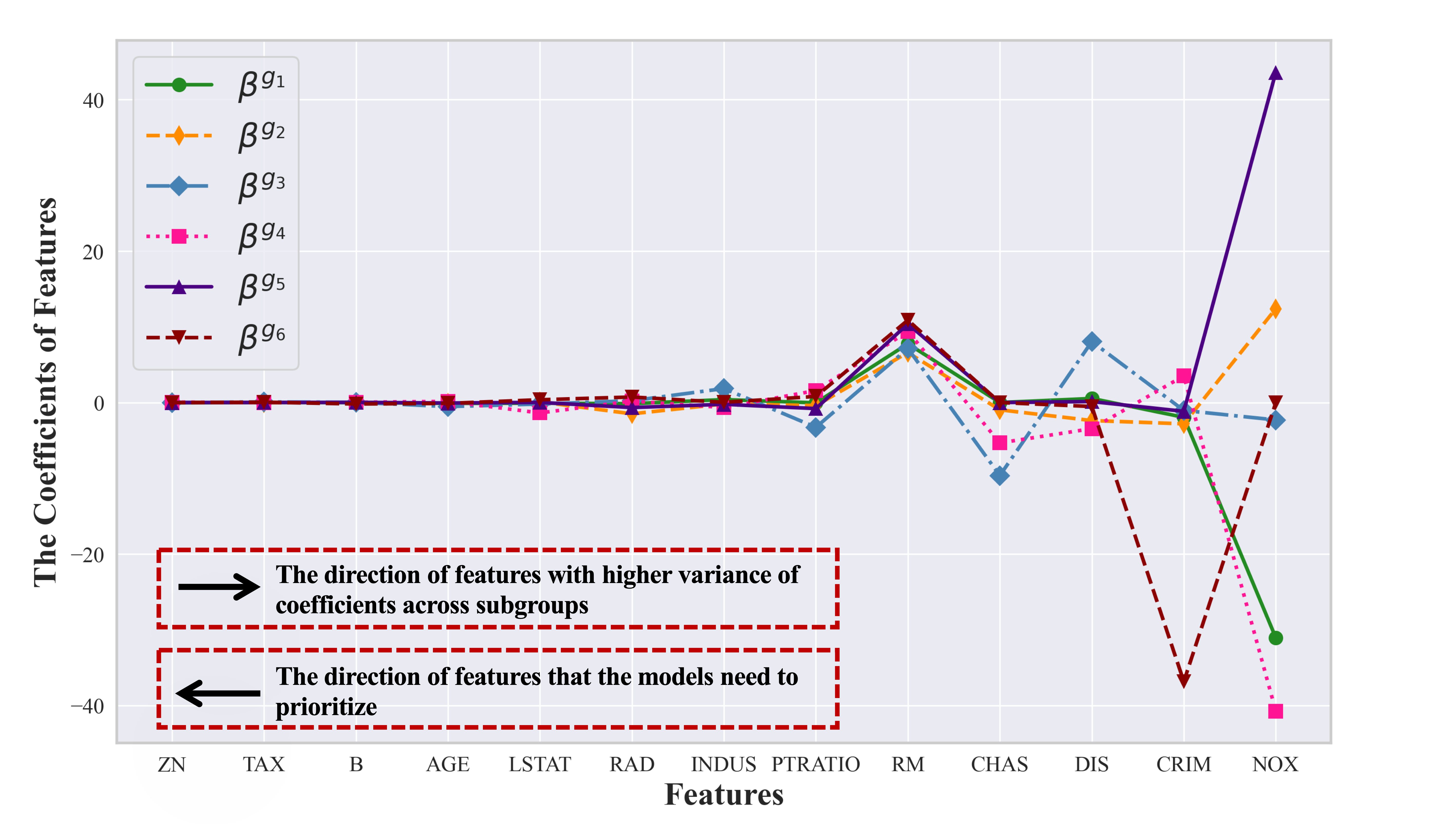}
    \caption{The coefficients $\beta^{\text{g}_{i}}$ of features across heterogeneous subgroups $G=\{\text{g}_{1},\text{g}_{2},\dots,\text{g}_{6}\}$ as learned by the method with $\text{K}=6$. The features with lower variances of coefficients are more likely causal factors that the models need to prioritize.}
    \label{fig3}
\end{figure}

For the heterogeneous subgroups identified by the method $\text{G}=\{\text{g}_{1},\text{g}_{2},\dots,\text{g}_{k}\}$, the regression coefficient $\beta_{p}^{\text{g}_{i}}$ for each feature $X_{p}$ within each identified subgroup $\text{g}_{i}$ is first calculated, representing the relationship between $X_{p}$ and target using traditional regression method:
\begin{equation}
Y^{\text{g}_{i}}=\beta_{0}^{\text{g}_{i}}+\beta_{1}^{\text{g}_{i}}X_{1}+\cdots+\beta_{m}^{\text{g}_{i}}X_{m}.
\end{equation}
By aggregating the coefficients across different subgroups, each feature can be transformed into a vector space representing the combination of relationships under heterogeneous subgroups:
\begin{equation}
    \text{F}(X_{p})=(\beta_{p}^{\text{g}_{1}},\beta_{p}^{\text{g}_{2}},\dots,\beta_{p}^{\text{g}_{k}}).
\end{equation}
The lower variance between the vector elements for the feature indicates a more stable relationship with the target. Thus, the feature is more likely a causal factor that the models need to prioritize.

\begin{table*}[h]
\renewcommand\arraystretch{1.25}
\centering
\scriptsize
\caption{The results of the method after deploying to popular causal learning methods across various cluster scenarios $\text{K}=\{2,3,4,5,6\}$, where the best results are denoted as bolded. The method is robust to the selection of parameter $\text{K}$, reducing the predicted error in almost all scenarios (underlines).}
\begin{tabular}{|c|cc|cc|cc|cc|cc|}
\hline
Methods                            & \multicolumn{2}{c|}{ERM \cite{r16}}                        & \multicolumn{2}{c|}{KerHRM \cite{r17}}                       & \multicolumn{2}{c|}{DWR \cite{DWR}}                         & \multicolumn{2}{c|}{SRDO \cite{SRDO}}                       & \multicolumn{2}{c|}{SVI \cite{SVI}}                         \\ \hline
Cluster Scenarios & \multicolumn{1}{c|}{RMSE} & Sta\_Error          & \multicolumn{1}{c|}{RMSE}  & Sta\_Error           & \multicolumn{1}{c|}{RMSE} & Sta\_Error           & \multicolumn{1}{c|}{RMSE} & Sta\_Error          & \multicolumn{1}{c|}{RMSE} & Sta\_Error           \\ \cline{2-11} 
                                   & \multicolumn{1}{c|}{3.09} & 24.68               & \multicolumn{1}{c|}{21.43} & 23.43                & \multicolumn{1}{c|}{0.80} & 26.57                & \multicolumn{1}{c|}{0.22} & 19.11               & \multicolumn{1}{c|}{0.41} & 15.30                \\ \hline
+CKC(K=2)                          & \multicolumn{1}{c|}{4.70} & { \textbf{\underline{8.60}}} & \multicolumn{1}{c|}{18.92} & { \underline{11.41}}          & \multicolumn{1}{c|}{1.96} & { \textbf{\underline{13.46}}} & \multicolumn{1}{c|}{0.48} & { \underline{7.68}}          & \multicolumn{1}{c|}{0.56} & { \textbf{\underline{12.75}}} \\ \hline
+CKC(K=3)                          & \multicolumn{1}{c|}{3.29} & { \underline{8.83}}          & \multicolumn{1}{c|}{12.31} & { \textbf{\underline{10.54}}} & \multicolumn{1}{c|}{1.56} & { \underline{14.30}}          & \multicolumn{1}{c|}{0.23} & { \underline{8.60}}          & \multicolumn{1}{c|}{0.45} & { \underline{12.99}}          \\ \hline
+CKC(K=4)                          & \multicolumn{1}{c|}{4.69} & { \underline{12.23}}         & \multicolumn{1}{c|}{24.03} & { \underline{12.01}}          & \multicolumn{1}{c|}{1.95} & { \underline{19.46}}          & \multicolumn{1}{c|}{0.67} & { \underline{13.92}}         & \multicolumn{1}{c|}{0.45} & { \underline{13.63}}          \\ \hline
+CKC(K=5)                          & \multicolumn{1}{c|}{4.68} & { \underline{12.39}}         & \multicolumn{1}{c|}{10.92} & { \underline{13.86}}          & \multicolumn{1}{c|}{1.92} & { \underline{26.11}}          & \multicolumn{1}{c|}{0.44} & { \textbf{\underline{6.88}}} & \multicolumn{1}{c|}{0.44} & { \underline{13.35}}          \\ \hline
+CKC(K=6)                          & \multicolumn{1}{c|}{3.29} & { \underline{8.82}}          & \multicolumn{1}{c|}{22.26} & { \underline{11.78}}          & \multicolumn{1}{c|}{1.54} & { \underline{20.39}}          & \multicolumn{1}{c|}{0.25} & { \underline{9.07}}          & \multicolumn{1}{c|}{0.45} & { \underline{13.25}}          \\ \hline
\end{tabular}
\label{tab3}
\end{table*}

\textit{Results.} Fig. \ref{fig3} displays the coefficients of features under different subgroups $\text{G}=\{\text{g}_{1},\text{g}_{2},\dots,\text{g}_{6}\}$ as learned by the method with $\text{K}=6$. This choice of $\text{K}$ facilitates a clearer and more distinct presentation of the results. The method effectively utilizes heterogeneous subgroups to identify the causal factors, directing models to prioritize them. Table \ref{tab3} outlines the outcomes of deploying the method to popular causal learning methods across diverse cluster scenarios $\text{K}=\{2,3,4,5,6\}$, demonstrating the effectiveness of the method in enhancing causal learning. An important observation is the robustness of the method across varying cluster scenarios $\text{K}$. This suggests that determining the precise number of heterogeneous subgroups in advance is not crucial, as long as it accurately captures the heterogeneous subgroup information. The flexibility embedded in the method enhances its generalizability, enabling adaptation to various scenarios without compromising its effectiveness in capturing heterogeneous subgroups.

\section{Conclusion}
The challenge addressed in this work pertains to heterogeneous subgroup causal learning in the context of multi-source and heterogeneous data. This issue arises from the need to account for variations in causal relationships that a single causal model may not fully capture. To tackle this, a method is considered for heterogeneous subgroup causal learning through the nonlinear \underline{\textbf{C}}ausal \underline{\textbf{K}}ernel \underline{\textbf{C}}lustering (CKC). Diverging from conventional methods that apply a single causal model to the entire data set, often resulting in insufficient representation and overlooking data diversity nuances, the method introduces a $u$-centered based sample mapping function and a corresponding nonlinear causal kernel for clustering. This ensures that samples within each cluster exhibit more closely aligned causal relationships. Serving as a flexible causal discovery module, the method captures heterogeneous subgroups across different data types while enhancing causal learning, \textcolor{new}{making it a valuable tool for various applications involving multi-source and heterogeneous data.}

\bibliographystyle{unsrt}  
\bibliography{references}  

\begin{thebibliography}{10}

\bibitem{r1}
Clark Glymour, Kun Zhang, and Peter Spirtes.
\newblock {Review of Causal Discovery Methods Based on Graphical Models}.
\newblock {\em Frontiers in Genetics}, 10, 2019.

\bibitem{r2}
Bernhard Schölkopf, Francesco Locatello, Stefan Bauer, Nan~Rosemary Ke, Nal Kalchbrenner, Anirudh Goyal, and Yoshua Bengio.
\newblock {Toward Causal Representation Learning}.
\newblock {\em Proceedings of the IEEE}, 109(5):612--634, 2021.

\bibitem{others2}
Alex Markham and Moritz Grosse-Wentrup.
\newblock {Measurement Dependence Inducing Latent Causal Models}.
\newblock In {\em Proceedings of the 36th Conference on Uncertainty in Artificial Intelligence (UAI)}, pages 590--599. PMLR, 2020.

\bibitem{others5}
Paul~K Rubenstein, Sebastian Weichwald, Stephan Bongers, Joris~M Mooij, Dominik Janzing, Moritz Grosse-Wentrup, and Bernhard Sch{\"o}lkopf.
\newblock {Causal Consistency of Structural Equation Models}.
\newblock {\em arXiv preprint arXiv:1707.00819}, 2017.

\bibitem{r3}
Yunxia Wang, Fuyuan Cao, Kui Yu, and Jiye Liang.
\newblock {Local Causal Discovery in Multiple Manipulated Datasets}.
\newblock {\em IEEE Transactions on Neural Networks and Learning Systems}, 34(10):7235--7247, 2023.

\bibitem{r4}
Ruocheng Guo, Lu~Cheng, Jundong Li, P.~Richard Hahn, and Huan Liu.
\newblock {A Survey of Learning Causality with Data: Problems and Methods}.
\newblock {\em ACM Comput. Surv.}, 53(4), jul 2020.

\bibitem{r6}
Biwei Huang, Kun Zhang, Jiji Zhang, Joseph Ramsey, Ruben Sanchez-Romero, Clark Glymour, and Bernhard Schölkopf.
\newblock {Causal Discovery from Heterogeneous/Nonstationary Data}.
\newblock {\em Journal of Machine Learning Research}, 21(89):1--53, 2020.

\bibitem{r19}
Georgios Mavroudeas, Nafis Neehal, Jason Kuruzovich, Kristin~P. Bennett, and Malik Magdon-Ismail.
\newblock {Subpopulation Analysis in Causal Inference: A Healthcare Case Study}.
\newblock In {\em 2022 IEEE International Conference on Bioinformatics and Biomedicine (BIBM)}, pages 1673--1676, 2022.

\bibitem{r20}
Wei Chen, Yunjin Wu, Ruichu Cai, Yueguo Chen, and Zhifeng Hao.
\newblock {CCSL: a causal structure learning method from multiple unknown environments}.
\newblock {\em arXiv preprint arXiv:2111.09666}, 2021.

\bibitem{ker2}
Alex Markham, Richeek Das, and Moritz Grosse-Wentrup.
\newblock {A Distance Covariance-based Kernel for Nonlinear Causal Clustering in Heterogeneous Populations}.
\newblock In {\em Proceedings of the First Conference on Causal Learning and Reasoning}, volume 177 of {\em Proceedings of Machine Learning Research}, pages 542--558. PMLR, 2022.

\bibitem{r7}
Vaishali Mahipal and Mohammad Arif~Ul Alam.
\newblock {Estimating Heterogeneous Causal Effect of Polysubstance Usage on Drug Overdose from Large-Scale Electronic Health Record}.
\newblock In {\em 2022 44th Annual International Conference of the IEEE Engineering in Medicine {\&} Biology Society (EMBC)}, pages 1028--1031, 2022.

\bibitem{r17}
Jiashuo Liu, Zheyuan Hu, Peng Cui, Bo~Li, and Zheyan Shen.
\newblock {Integrated Latent Heterogeneity and Invariance Learning in Kernel Space}.
\newblock In {\em {Advances in Neural Information Processing Systems}}, volume~34, pages 21720--21731. Curran Associates, Inc., 2021.

\bibitem{r10}
Sindy L{\"o}we, David Madras, Richard Zemel, and Max Welling.
\newblock {Amortized Causal Discovery: Learning to Infer Causal Graphs from Time-Series Data}.
\newblock In {\em Proceedings of the First Conference on Causal Learning and Reasoning}, volume 177 of {\em Proceedings of Machine Learning Research}, pages 509--525. PMLR, 11--13 Apr 2022.

\bibitem{r11}
Chenxi Liu and Kun Kuang.
\newblock {Causal Structure Learning for Latent Intervened Non-stationary Data}.
\newblock In {\em Proceedings of the 40th International Conference on Machine Learning}, volume 202 of {\em Proceedings of Machine Learning Research}, pages 21756--21777. PMLR, 23--29 Jul 2023.

\bibitem{r13}
Shoubo Hu, Zhitang Chen, Vahid Partovi~Nia, Laiwan CHAN, and Yanhui Geng.
\newblock {Causal Inference and Mechanism Clustering of A Mixture of Additive Noise Models}.
\newblock In {\em Advances in Neural Information Processing Systems}, volume~31. Curran Associates, Inc., 2018.

\bibitem{r14}
Xingyu Wu, Bingbing Jiang, Yan Zhong, and Huanhuan Chen.
\newblock {Multi-label causal variable discovery: Learning common causal variables and label-specific causal variables}.
\newblock {\em arXiv preprint arXiv:2011.04176}, 2020.

\bibitem{r15}
Abdellah Rahmani and Pascal Frossard.
\newblock {Causal Temporal Regime Structure Learning}.
\newblock {\em arXiv preprint arXiv:2311.01412}, 2024.

\bibitem{s1}
Matthew~J. Vowels, Necati~Cihan Camgoz, and Richard Bowden.
\newblock {D’ya Like DAGs? A Survey on Structure Learning and Causal Discovery}.
\newblock {\em ACM Comput. Surv.}, 55(4), November 2022.

\bibitem{lemma1}
G{\'a}bor~J. Sz{\'e}kely and Maria~L. Rizzo.
\newblock {Partial distance correlation with methods for dissimilarities}.
\newblock {\em The Annals of Statistics}, 42(6):2382 -- 2412, 2014.

\bibitem{r21}
Elias Bareinboim and Judea Pearl.
\newblock {Causal inference and the data-fusion problem}.
\newblock {\em Proceedings of the National Academy of Sciences}, 113(27):7345--7352, 2016.

\bibitem{r22}
Ankit Sharma, Garima Gupta, Ranjitha Prasad, Arnab Chatterjee, Lovekesh Vig, and Gautam Shroff.
\newblock {MetaCI: Meta-learning for causal inference in a heterogeneous population}.
\newblock {\em arXiv preprint arXiv:1912.03960}, 2019.

\bibitem{r23}
Ahmed~M. Alaa and Mihaela van~der Schaar.
\newblock {Bayesian Inference of Individualized Treatment Effects using Multi-task Gaussian Processes}.
\newblock In {\em Advances in Neural Information Processing Systems}, volume~30. Curran Associates, Inc., 2017.

\bibitem{DAG1}
Ignavier Ng, Shengyu Zhu, Zhuangyan Fang, Haoyang Li, Zhitang Chen, and Jun Wang.
\newblock Masked gradient-based causal structure learning.
\newblock In {\em Proceedings of the 2022 SIAM International Conference on Data Mining (SDM)}, pages 424--432. SIAM, 2022.

\bibitem{r27}
Dag Tj{\o}stheim, H{\aa}kon Otneim, and B{\aa}rd St{\o}ve.
\newblock {Statistical Dependence: Beyond Pearson’s $\rho$}.
\newblock {\em Statistical Science}, 37(1):90--109, 2022.

\bibitem{r28}
Shun Yao, Xianyang Zhang, and Xiaofeng Shao.
\newblock {Testing Mutual Independence in High Dimension via Distance Covariance}.
\newblock {\em Journal of the Royal Statistical Society Series B: Statistical Methodology}, 80(3):455--480, 10 2017.

\bibitem{lemma2}
Arthur Gretton, Kenji Fukumizu, and Bharath~K Sriperumbudur.
\newblock {Discussion of: Brownian distance covariance}.
\newblock {\em The annals of applied statistics}, 3(4):1285--1294, 2009.

\bibitem{tj1}
Mary~L McHugh.
\newblock The {C}hi-square test of independence.
\newblock {\em Biochemia Medica}, 23(2):143--149, 2013.

\bibitem{tj2}
Eric Benhamou and Valentin Melot.
\newblock Seven proofs of the {P}earson {C}hi-squared independence test and its graphical interpretation.
\newblock {\em arXiv preprint arXiv:1808.09171}, 2018.

\bibitem{tj3}
Sambit~Panda Cencheng~Shen and Joshua~T. Vogelstein.
\newblock {The Chi-Square Test of Distance Correlation}.
\newblock {\em Journal of Computational and Graphical Statistics}, 31(1):254--262, 2022.
\newblock PMID: 35707063.

\bibitem{corr2}
Khawla Ali Abd Al-Hameed.
\newblock Spearman's correlation coefficient in statistical analysis.
\newblock {\em International Journal of Nonlinear Analysis and Applications}, 13(1):3249--3255, 2022.

\bibitem{discorr2}
Shubhadeep Chakraborty and Xianyang Zhang.
\newblock {A new framework for distance and kernel-based metrics in high dimensions}.
\newblock {\em Electronic Journal of Statistics}, 15(2):5455 -- 5522, 2021.

\bibitem{linear2}
Feng Xie, Biwei Huang, Zhengming Chen, Yangbo He, Zhi Geng, and Kun Zhang.
\newblock {Identification of Linear Non-{G}aussian Latent Hierarchical Structure}.
\newblock In {\em Proceedings of the 39th International Conference on Machine Learning}, volume 162 of {\em Proceedings of Machine Learning Research}, pages 24370--24387. PMLR, 17--23 Jul 2022.

\bibitem{linear1}
Jeffrey Adams, Niels Hansen, and Kun Zhang.
\newblock {Identification of Partially Observed Linear Causal Models: Graphical Conditions for the Non-Gaussian and Heterogeneous Cases}.
\newblock In {\em Advances in Neural Information Processing Systems}, volume~34, pages 22822--22833. Curran Associates, Inc., 2021.

\bibitem{freedom1}
Amy Nowacki.
\newblock {Chi-square and Fisher’s exact tests}.
\newblock {\em Cleve Clin J Med}, 84(9 suppl 2):e20--5, 2017.

\bibitem{freedom2}
Weddha Savitri, Ni~Luh~Sutjiati Beratha, I~Nengah Sudipa, and I~Made Rajeg.
\newblock {Applying Chi-Square Test In Measuring The Significance Of The Occurrence Of French Synonym In Corpus Data}.
\newblock {\em International Journal of Linguistics and Discourse Analytics}, 6(1):13--21, 2024.

\bibitem{ker3}
Fateme~Nateghi Haredasht, Farnaz Ghassemi, and Mohammad~Hassan Moradi.
\newblock {Causal inference of gene expression data using a clustering-based extension of Kernel-Granger causality}.
\newblock In {\em 2016 23rd Iranian Conference on Biomedical Engineering and 2016 1st International Iranian Conference on Biomedical Engineering (ICBME)}, pages 84--88, 2016.

\bibitem{s2}
Yuhao Wang, Santiago Segarra, and Caroline Uhler.
\newblock {High-dimensional joint estimation of multiple directed Gaussian graphical models}.
\newblock {\em Electronic Journal of Statistics}, 14(1):2439 -- 2483, 2020.

\bibitem{Fcausal1}
Xingxuan Zhang, Peng Cui, Renzhe Xu, Linjun Zhou, Yue He, and Zheyan Shen.
\newblock {Deep Stable Learning for Out-of-Distribution Generalization}.
\newblock In {\em Proceedings of the IEEE/CVF Conference on Computer Vision and Pattern Recognition (CVPR)}, pages 5372--5382, June 2021.

\bibitem{F1}
Jiaxuan Liang, Jun Wang, Guoxian Yu, Shuyin Xia, and Guoyin Wang.
\newblock {Multi-Granularity Causal Structure Learning}.
\newblock {\em Proceedings of the AAAI Conference on Artificial Intelligence}, 38(12):13727--13735, Mar. 2024.

\bibitem{tonggou1}
Thomas~W Judson.
\newblock {\em Abstract algebra: theory and applications}.
\newblock 2020.

\bibitem{tonggou2}
Sandhya~S Pai and Thankanchan Baiju.
\newblock {Continuous mappings in soft lattice topological spaces}.
\newblock {\em Italian J. Pure Appl. Math.(Communicated)}, 48:937--949, 2022.

\bibitem{ys2}
Kacper Pluta, Pascal Romon, Yukiko Kenmochi, and Nicolas Passat.
\newblock Bijective digitized rigid motions on subsets of the plane.
\newblock {\em Journal of Mathematical Imaging and Vision}, 59:84--105, 2017.

\bibitem{ys3}
Jiří Fiala and Jan Kratochvíl.
\newblock {Locally constrained graph homomorphisms—structure, complexity, and applications}.
\newblock {\em Computer Science Review}, 2(2):97--111, 2008.

\bibitem{tonggou3}
Xiaorui Sun.
\newblock {Faster Isomorphism for $p$-Groups of Class 2 and Exponent $p$}.
\newblock In {\em Proceedings of the 55th Annual ACM Symposium on Theory of Computing}, STOC 2023, page 433–440, New York, NY, USA, 2023. Association for Computing Machinery.

\bibitem{others1}
Alex Markham, Danai Deligeorgaki, Pratik Misra, and Liam Solus.
\newblock {A Transformational Characterization of Unconditionally Equivalent Bayesian Networks}.
\newblock In {\em Proceedings of The 11th International Conference on Probabilistic Graphical Models}, pages 109--120. PMLR, 2022.

\bibitem{matrix1}
Zhaobin Mo, Qingyuan Liu, Baohua Yan, Longxiang Zhang, and Xuan Di.
\newblock {Causal Adjacency Learning for Spatiotemporal Prediction Over Graphs}.
\newblock {\em arXiv preprint arXiv:2411.16142}, 2024.

\bibitem{r29}
Zhenghui Lu, Wenjie Dong, Bo~Lu, Naiming Yuan, Zhuguo Ma, Mikhail~I. Bogachev, and Juergen Kurths.
\newblock {Early warning of the Indian Ocean Dipole using climate network analysis}.
\newblock {\em Proceedings of the National Academy of Sciences}, 119(11):e2109089119, 2022.

\bibitem{r30}
Kristina~P. Sinaga and Miin-Shen Yang.
\newblock {Unsupervised K-Means Clustering Algorithm}.
\newblock {\em IEEE Access}, 8:80716--80727, 2020.

\bibitem{r31}
Vahid {Hooshmand Moghaddam} and Javad Hamidzadeh.
\newblock {New Hermite orthogonal polynomial kernel and combined kernels in Support Vector Machine classifier}.
\newblock {\em Pattern Recognition}, 60:921--935, 2016.

\bibitem{r32}
Zuzana Majdisova and Vaclav Skala.
\newblock Radial basis function approximations: comparison and applications.
\newblock {\em Applied Mathematical Modelling}, 51:728--743, 2017.

\bibitem{s5}
Ari~M. Lipsky and Sander Greenland.
\newblock {Causal Directed Acyclic Graphs}.
\newblock {\em JAMA}, 327(11):1083--1084, 03 2022.

\bibitem{s4}
G.~Di~Capua, J.~Runge, R.~V. Donner, B.~van~den Hurk, A.~G. Turner, R.~Vellore, R.~Krishnan, and D.~Coumou.
\newblock {Dominant patterns of interaction between the tropics and mid-latitudes in boreal summer: causal relationships and the role of timescales}.
\newblock {\em Weather and Climate Dynamics}, 1(2):519--539, 2020.

\bibitem{r16}
V.~Vapnik.
\newblock {Principles of Risk Minimization for Learning Theory}.
\newblock In {\em Advances in Neural Information Processing Systems}, volume~4. Morgan-Kaufmann, 1991.

\bibitem{DWR}
Kun Kuang, Ruoxuan Xiong, Peng Cui, Susan Athey, and Bo~Li.
\newblock {Stable Prediction with Model Misspecification and Agnostic Distribution Shift}.
\newblock {\em Proceedings of the AAAI Conference on Artificial Intelligence}, 34(04):4485--4492, Apr. 2020.

\bibitem{SRDO}
Zheyan Shen, Peng Cui, Tong Zhang, and Kun Kunag.
\newblock {Stable Learning via Sample Reweighting}.
\newblock {\em Proceedings of the AAAI Conference on Artificial Intelligence}, 34(04):5692--5699, Apr. 2020.

\bibitem{SVI}
Han Yu, Peng Cui, Yue He, Zheyan Shen, Yong Lin, Renzhe Xu, and Xingxuan Zhang.
\newblock {Stable Learning via Sparse Variable Independence}.
\newblock {\em Proceedings of the AAAI Conference on Artificial Intelligence}, 37(9):10998--11006, Jun. 2023.

\end{thebibliography}
\end{document}